\documentclass[journal]{IEEEtran}
\IEEEoverridecommandlockouts
\usepackage{cite}
\usepackage{amsmath,amssymb,amsfonts}
\usepackage{algorithmic}
\usepackage{graphics}
\usepackage[ruled,vlined]{algorithm2e}
\usepackage{graphicx}
\usepackage{textcomp}
\usepackage{xcolor}
\usepackage{soul}
\usepackage{multirow}
\usepackage{verbatim}
\usepackage{subfig}
\usepackage{caption} 
\usepackage{longtable}
\usepackage{url}

\def\BibTeX{{\rm B\kern-.05em{\sc i\kern-.025em b}\kern-.08em
    T\kern-.1667em\lower.7ex\hbox{E}\kern-.125emX}}

\begin{document}

\title{\vspace{6mm} 
Only Pick Once -- Multi-Object Picking Algorithms for Picking Exact Number of Objects Efficiently
}
\author{Zihe Ye and Yu Sun
\thanks{*This material is based upon work supported by the National Science Foundation under Grants Nos. 1812933 and 191004.}
\thanks{The authors are from the Robot Perception and Action Lab (RPAL) of Computer Science and Engineering Department, University of South Florida, Tampa, FL 33620, USA. Email: \texttt{\{ziheye,yusun\}@usf.edu}.}%
}

\maketitle

\begin{abstract}
Picking up multiple objects at once is a grasping skill that makes a human worker efficient in many domains. This paper presents a system to pick a requested number of objects by only picking once (OPO). The proposed Only-Pick-Once System (OPOS) contains several graph-based algorithms that convert the layout of objects into a graph, cluster nodes in the graph, rank and select candidate clusters based on their topology. OPOS also has a multi-object picking predictor based on a convolutional neural network for estimating how many objects would be picked up with a given gripper location and orientation. This paper presents four evaluation metrics and three protocols to evaluate the proposed OPOS. The results show OPOS has very high success rates for two and three objects when only picking once. Using OPOS can significantly outperform two to three times single object picking in terms of efficiency. The results also show OPOS can generalize to unseen size and shape objects. 

\end{abstract}

\section{INTRODUCTION}

\begin{figure}[h!]
    \centering
    \includegraphics[width=0.7\linewidth]{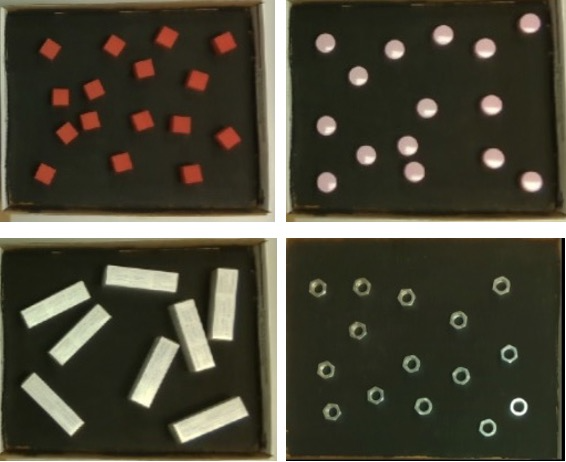}
    \caption{Examples scenes of batch picking for four shapes: cube, cylinder, cuboid, hexagon.}
    \label{target_scene}
\end{figure}

In warehouses, workers usually perform batch picking to improve efficiency, also called multi-order picking. It is picking several same objects from a bin at once for multiple orders. For instance, a worker could be instructed to pick four boxes of toothpaste or three jars of a cosmetic product from a bin. In manufacturing, when putting nuts on a set of screws, workers usually get several nuts by only picking once (OPO) and then put them on one by one. During food prep, we usually pick multiple chunks from a cutting board at once and drop them into a pan after we chop vegetables into identical chunks. Figure \ref{target_scene} shows four examples of multiple objects randomly lying in a bin. 


People pick up several objects at once instead of picking single object for multiple times for efficiency. For example, if a robotic system can pick up only one item and drop it into a bin in $3$ seconds, to pick up two identical items, the robot would need to pick two times, which is $6$ seconds. In contrast, a human worker can get the same two items by OPO and needs only $3$ seconds. That is why human workers are much faster than a robot. 

Efficient and reliable robot picking systems are in urgent demand to relieve recent labor shortage issue\cite{danielczuk2018linear, mahler2017learning,matsumura2019tangle,agrawal2010vision}. Numerous robot-picking systems have been developed for bin-picking, mainly applying single-object picking (SOP) strategies since the main research and development focus on SOP. Several early works on grasping stability analysis provide fundamental investigations on the mechanism of holding multiple objects \cite{yamada2009planar,yamada2012multifinger, yamada2005friction,yamada2005grasp,yamada2015static}. Studies on force closure and kinematics of multiple objects grasping have also been performed in  \cite{harada1998envoloping, harada1998kinematics,harada2000active,harada2000neighbor,harada2000rolling}. Nevertheless, none of these works studied how to pick multiple objects.
Only lately, the concept of picking up multiple objects started to emerge in response to the demand for robots to match human-level efficiency. Recently, researchers started looking into how to pick multiple objects for different goals under different settings\cite{chen2021multie, Shenoy2022Multi, agboh2022multi,sakamoto2021efficient, agboh2022learning}. 

This work tackles the problem of getting a requested number of identical objects in a shallow bin by OPO using a simple parallel gripper. It compensates for our previous work on multiple-object grasping (MOG) by OPO, since our previous work focused on piled-up objects in a bin \cite{Shenoy2022Multi} with a dexterous robotic hand. The previous work cannot handle the scenario where all objects lie un-stacked on the bottom of the bin. However, the un-stacked scenario requires more attention since it is very common in e-commerce warehouses and flexible manufacturing where stocks are kept low. 

To pick up multiple objects efficiently, we develop the Only-Pick-Once system (OPOS) that models the relationship among the objects as a graph. The algorithms in OPOS first group nearby objects into local clusters based on their graph representation and rank them based on their clique orders and connections with out-clique nodes. OPOS also has a multi-object picking (MOP) module guided by OPO predictor to estimate the outcome of picking a cluster of objects based on their layouts. With it, OPOS can select the best collision-free picking pose from a few candidates calculated based on the cluster's layout for a requested number of objects by only picking once. OPOS also includes algorithms balancing the computational cost and success rate for complex layouts to make the approach feasible to run in real applications. 


We have designed four evaluation metrics and protocols to rigorously and thoroughly evaluate the proposed approach. The approach is evaluated in both a simulation environment and a real setup. The evaluation results show that our approach achieves high success rates matching the requested number when only picking once. It leads to significantly reduced number of picking actions when using it in batch picking compared to the single-object picking strategy. The evaluation has also demonstrated that the approach can generalize well to other unseen objects with different sizes and shapes. 


Overall this paper has the following contributions: 
\begin{itemize}
    \item as far as we know, this paper is the first attempt to provide a solution for picking up the requested number of objects by only picking once; 
    \item this paper proposes a novel and comprehensive system to solve the OPO problem and achieves much better efficiency than SOP while ensuring picking accuracy; 
    \item this paper models objects in a graph and applies several graph algorithms to analyze the object layouts for multi-object picking;
    \item this paper presents a cluster ranking algorithm to significantly reduce the computation cost in searching for the right cluster;
    \item this paper presents a MOP predictor that can estimate the picking outcomes based on the cluster layout;
    \item for evaluation, the paper defines novel metrics and protocols to measure the performance of OPOS;
    \item this paper presents a rigorous and thorough evaluation and results of the proposed approach in a simulation and the real world;
\end{itemize}


\section{Related Works}
\label{related_works}
\subsection{Single Object Picking}
\label{single_related}
Single object bin-picking has been a topic of significant research for several decades. It usually uses a vision system to estimate an object's pose given its model or directly find grasp points. Traditional approaches use the known object CAD model and features in 2D or 3D images to estimate the object's pose \cite{lowe1991fitting, dementhon1995model, zhu2014single} and then transform predefined grasps in the CAD coordinate into the application coordinate \cite{ikeuchi1983picking, agrawal2010vision}. Lately, with large labeled datasets and complicated prediction models (usually deep neural networks), researchers have developed approaches that can skip pose estimation and directly find proper grasp points from dense 3D point clouds \cite{mahler2017learning, 
lenz2015deep,ten2018using, kappler2015leveraging}. A comprehensive review can be found in \cite{li2019survey}.


\subsection{Stability and Force Closure Analysis of Multi-Object Grasping}
\label{stability_related}
Static grasp stability analysis of multiple objects with a robotic hand has also been investigated. \cite{harada1998envoloping, harada2000rolling} discuss the enveloping grasp of multiple objects under rolling contacts and studied force closure of multiple objects. It builds the theoretical basis for later work on active force closure analysis for the manipulation of multiple objects in \cite{harada2000active}. \cite{harada1998kinematics} studied kinematics and internal forces during multi-grasping process. \cite{harada2000neighbor} talks about neighborhood equilibrium in multiple object holding.
\cite{yoshikawa2001optimization, yamada2005grasp, yamada2015static, yamada2005friction} try to achieve stably grasping of multiple objects through force-closure-based strategies. None of these works consider how to pick up multiple objects.

\subsection{Multi-Object Picking in different settings}
\label{approach_to_multipick}
Recently several research groups have started to look into the multi-object grasping problem. Our previous work \cite{chen2021multie, Shenoy2022Multi} has modeled the multi-object grasping as a stochastic process for picking multiple objects from a pile in a bin. It can pick up the requested number of objects from a pile with a good success rate when only picking once. Our later work has also analyzed the grasp types and taxonomy exhibited during grasping and holding multiple objects \cite{sun2022multi}. 

For picking objects on a flat surface using a gripper, \cite{sakamoto2021efficient} uses a two-step dynamic programming approach to pick one or two objects depending on availability. For picking two objects, they either pick two objects at once or push-grasp two objects when possible, their goal is to minimize the movement length of robot arm end effector. \cite{agboh2022multi,agboh2022learning} propose a novel grasp planner to do multi-object picking to reduce number of picking actions, their approach aims to remove all objects on a table fast without requirement on the picking number each time. All works have similar workspace setting as ours, but their goal is to clean the entire workspace while our goal is to pick up the requested number of objects by only picking once. 

\cite{agboh2022learning} trains a neural network based on data collected in real world, the input to their neural network is the object cluster and picking pose information, while the neural network in our OPOS only uses the image of the area between the gripper fingers as its input. Our estimation model generalization gives robot the ability to pick up unseen objects with different sizes and shapes with high accuracy, and our training procedure can be used on different grippers.

\subsection{Multi-Object Manipulation}
Robot manipulation in cluttered environment is always research focus of robotics field. \cite{pan2022algorithms} does a survey about past works working on multiple object manipulation,  including singulation, navigation, declutter, rearrangement, packing, placing, sorting-by-packing, sorting-by-clustering.

\section{Problem Settings}
\label{proble_settings}

The objective is to develop a robotic solution for picking up multiple identical objects from a shallow bin by only picking once. The targeted scenarios include a shallow bin and several identical objects laying loosely in the bin without stacking. It is a common setup in many warehouses. The size and location of the bin are known to the robotic system and within the robotic system's workspace. The shapes and sizes of the objects can be different in different setups, and they are also known to the robotic system and reasonable to the robotic system's gripper limitations. In this paper, we assume the objects have regular shapes, such as cubes, cuboids, cylinders, and hexagons. 

There could be multiple goals when picking up numerous identical objects, such as picking up as many as possible, picking up the requested number, and picking up the requested number as efficiently as possible. This paper focuses on developing a solution for picking up the requested number of objects by only picking once. This paper also explores the efficiency benefit the OPOS could bring to batch picking. 

\section{Methodology}
\label{Methodology}
\subsection{Overview}
\begin{figure}[h]
    \centering
    \includegraphics[width=1\linewidth]{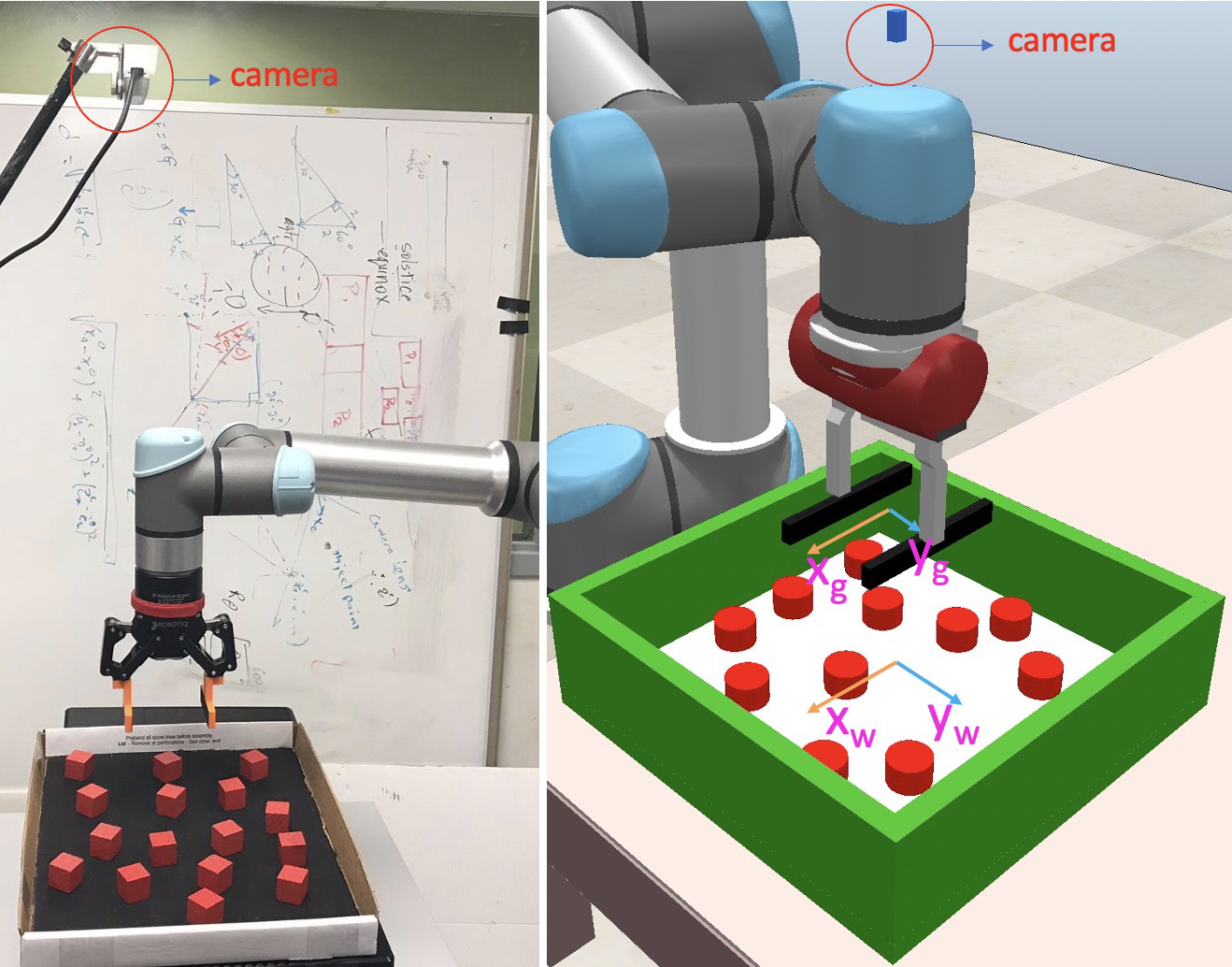}
    \caption{Robotic system with an RGB-D camera, a robot arm, and a parallel gripper. The system is for picking multiple objects from the shallow bin. The right figure shows gripper frame $x_g, y_g$ and world frame $x_w, y_w$. The coordinate frame between gripper and bin is identical in simulation and real setup.}
    \label{robot-setup}
\end{figure}

The proposed OPOS has a simple hand-eye robotic system. Figure \ref{robot-setup} shows our robotic system in real and CoppeliaSim. The system has an RGBD vision sensor (Intel RealSense Depth Camera D435), a 6-axis robotic arm (UR5), and a simple ``bang bang'' parallel gripper since logistics environments usually prefer low-cost pneumatic grippers. We use a Robotiq 2F-85 to simulate a ``bang bang'' gripper, only allowing full close and open. 

\begin{figure*}[htp!]
    \centering
    \vspace{-.cm}
    \includegraphics[width=\textwidth]{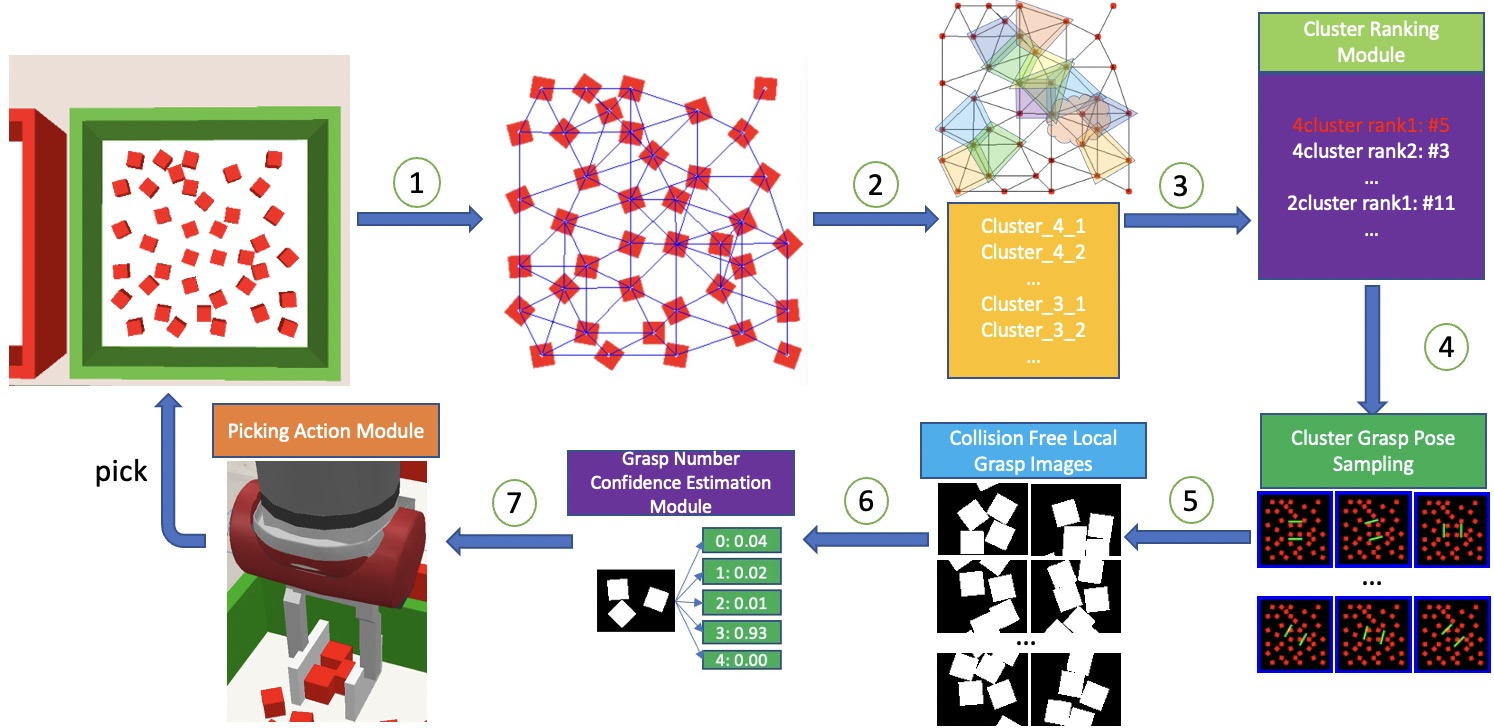}
    \caption{{\bf Overview} of the Only-Pick-Once System (OPOS). Module1 generates the neighbor graph. Module2 extracts clusters from the neighbor graph. Module3 ranks the clusters based on our ranking algorithm. Module4 generates grasping poses for the top-ranked cluster based on our sampling procedure. Module5 does collision checking on candidate grasping poses and keeps the collision free grasping poses. Module6 estimates the number of grasped objects with confidence for each grasping pose. Module7 checks the confidence with a preset threshold and passes the command to robot for action.}
    \vspace{-.cm}
    \label{high-level-flowchart}
\end{figure*}

Figure \ref{high-level-flowchart} provide an overview of the proposed OPOS. The input to the system is a single RGBD image - the top view of the bin, and the desired number $k$ of objects to be picked up. We assume $k$ is smaller than the max preset number $m$ based on the capability of the gripper. OPOS has a set of algorithms that process the image through seven modules and outputs a 2-DOF planar position $(x, y)$ and an in-plane rotation $\gamma$ to the robotic arm for picking. 

The module1 of OPOS is called \textit{neighbor graph generation}. In this module, an algorithm processes the RGBD image, extracts the objects from the image, obtains their locations, and generates a neighbor graph based on their relative positions. 
The module2 is called \textit{clustering}, in which clusters of $k$ to $m$ objects in the neighbor graph are identified. In module three, \textit{cluster ranking}, clusters are ranked based on a set of rules and stored in the ranked cluster list based on their ranking.

The first cluster in the ranked cluster list will be checked through the following three modules. If it is eliminated, the next cluster in the list becomes the top-ranked cluster and will be checked until the list runs out of clusters. 
In module4, \textit{picking pose proposal}, a sampling algorithm proposes several picking positions and orientations for the top-ranked cluster. 
In module5, \textit{collision checking}, the proposed picking poses are checked for collision. The collision-free poses are kept in a picking pose list. If there is no collision-free pose, the cluster is eliminated.
In module6, \textit{picking confidence estimation}, a trained neural network, \textit{multi-object picking predictor}, takes the local scene between the gripper fingers of a proposed pose in the picking pose list and estimates the confidence of picking up zero, one, two, and up to $m$ objects with that picking pose. 
In module7, \textit{picking pose selection}, the picking confidences of picking $k$ objects of various picking poses and clusters are compared with a pre-defined threshold and among themselves. The optimal picking pose is selected for execution.

Sections \ref{step1} to \ref{step7} provide detailed descriptions of the algorithms in the seven modules.

\subsection{Neighbor Graph Generation}
\label{step1} 
To pick up more than one object, we would need to inspect the relationships among objects and connect the ones the gripper could pick up together. Therefore, in the \textit{neighbor graph generation} module, our algorithm processes the input RGBD image, localizes the objects in the bin and generates a neighbor graph of objects based on their relative locations. So, the algorithm for this module has two parts: object center detection and graph generation. 

\begin{figure}[h!]
    \centering
    \includegraphics[width=1.0\linewidth]{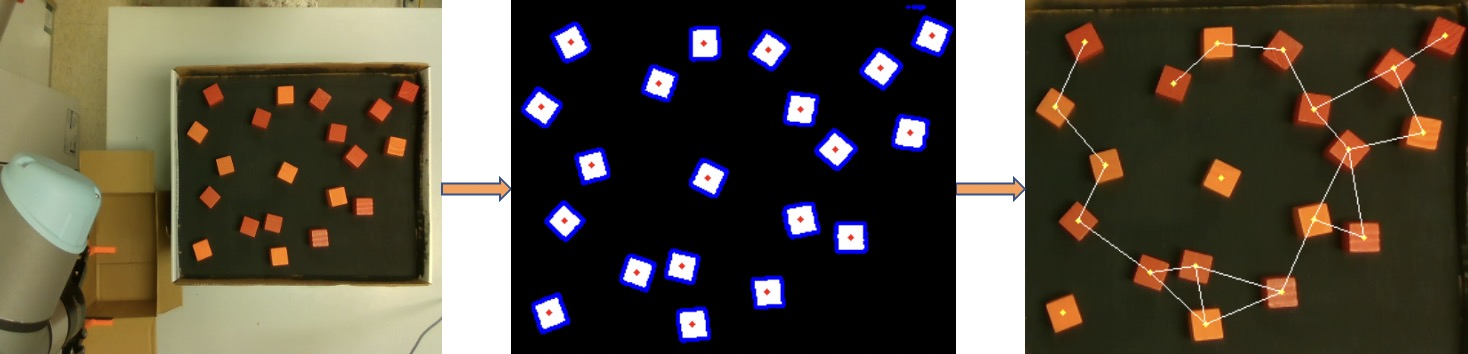}
    \caption{Neighbor graph generation example of 1inch wood-cube in real setup.}
    \label{contour_center_detection}
\end{figure}

An RGBD camera right above the bin takes an image of the bin with objects inside. Our algorithm first segments objects in the bin from the background, detects their contours and then estimates their centers. The algorithm then treats each object as a node and connects it with its neighbor nodes within a predefined distance. The neighborhood distance threshold $H_d$ is defined based on the gripper specifications (selection of $H_d$ is described in Section \ref{edge_threshold_definition}). Figure \ref{contour_center_detection} illustrates the two modules in the process using an example.

\begin{figure}[h!]
    \centering
    \includegraphics[height=0.325\linewidth]{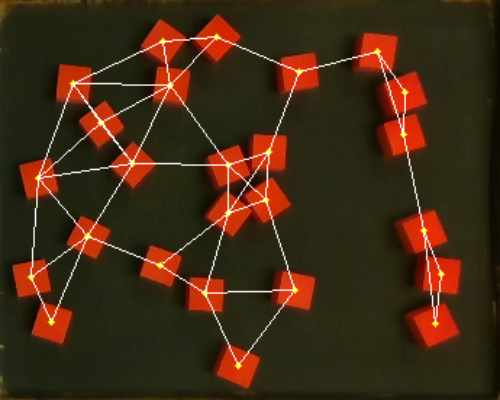}
    \includegraphics[height=0.325\linewidth]{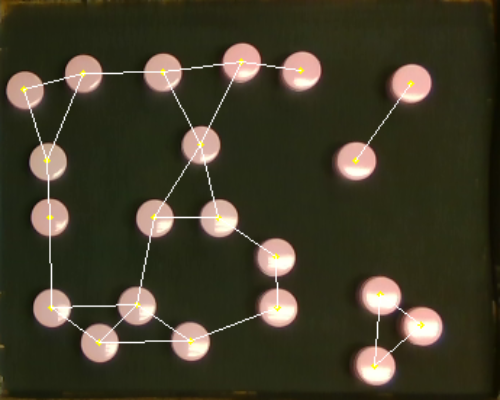}
    \caption{Neighbor graph example for 1inch Cube and 2.8cm Cylinder.}
    \label{cube_neighbor_graph}
\end{figure}

The final neighbor graph is an un-directed weighted graph $\bf{G} = \{N, E\}$. The nodes set $N$ contains the object indices and their location information. The edges set contains the connections and distance values. Figure \ref{cube_neighbor_graph} shows two more example neighbor graphs. In this module, the neighbor graph is generated, and the image of the bin area is normalized so that each pixel of the image represents a $1mm\times1mm$ area in the real space for next modules. 

\subsection{Clustering}
\label{step2} 
To pick up $k$ objects, we need to identify clusters with at least $k$ objects, the same as finding a k-clique in the neighbor graph. A $k$-clique means a $k$-complete graph where all $k$ nodes are fully connected. Therefore, our goal is to identify all cliques of $k$ order or higher in the neighbor graph.
We use the algorithm in the NetworkX library \cite{NetworkX} proposed in Zhang et al. \cite{clustering_algo} to find all cliques from one node clique to the max clique in the neighbor graph. The cliques with orders lower than $k$ are discarded, and the remaining are saved in the initial cluster list (ICL). 

The clique requirement is relatively loose; not every cluster in ICL could fit in the open gripper. First, we calculate the \textit{effective gripping area} of the open gripper. The width of the effective gripping area is defined as the gripper's open spread (distance between the gripper fingers). The length of the effective gripping area is defined as the length of the gripper finger plus one target object length (counting the half-target object length at both ends of the gripper finger). It is a generous definition since when the gripper closes, the object will usually slide out of the gripper if only a small portion of the object is in between the gripper fingers at the beginning. Therefore, if the center of an object is not in the effective gripping area, the object cannot be picked up by the gripper. 

So our algorithm \textbf{Algorithm  \ref{alg:clusterfit}} checks if the all objects of each cluster in ICL can fit in an effective gripping area. For each cluster, It first calculates the convex hull of all objects in the cluster, then uses the convex hull points to calculate a minimal area rectangle to enclose the entire convex hull \cite{min_area_rec}. We call this rectangle \textit{cluster rectangle}. Min\_Area\_Rec subroutine takes in a cluster and outputs the cluster rectangle. Rec\_in\_Rec subroutine takes in two arguments and checks if first rectangle can fit in the second rectangle. 

\begin{figure}[h!]
    \centering
    \includegraphics[width=1.0\linewidth]{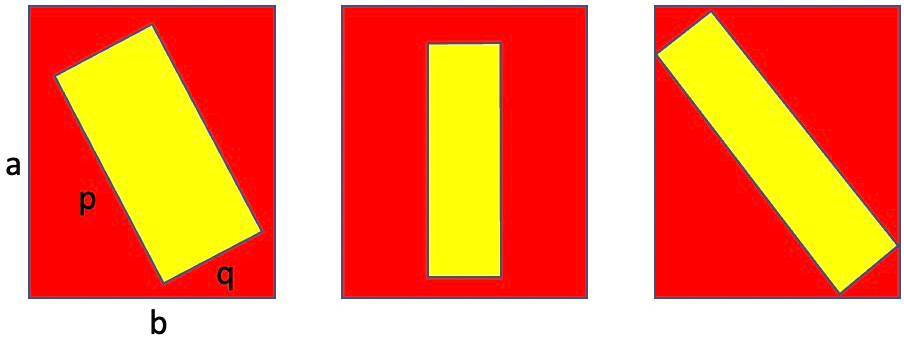}
    \\
   (A)~~~~~~~~~~~~~~~~~~~~~~~(B)~~~~~~~~~~~~~~~~~~~~~~(C)
    \caption{(A) a general scenario with one inner rectangle \textbf{M} whose longer side length is p, shorter side length is q, and one outer rectangle \textbf{G} whose longer side length is a, shorter side length is b. (B) and (C) are two example cases, both cases $a=10.0cm, b=8.4cm$. Part B: $p=8.0cm, q=2.5cm$. Part C: $p=10.25cm, q=2.5cm$.}
    \label{fig-rec-in-rec}
\end{figure}

To test if the cluster rectangle can fit in the effective gripper area, we use the well-known rec-in-rec theory \cite{rec_in_rec}. According to the rec-in-rec theory, as illustrated in Figure \ref{fig-rec-in-rec}(A), the yellow rectangle ($p \times q$ area, $q \leq p$) can fit in the red rectangle ($a \times b$ area, $b \leq a$) if and only if either of the following conditions is true:

(a) $p \leq a$ \textit{and} $q \leq b$

(b) $p > a$, $q \leq b$, \textit{and} $(\frac{a+b}{p+q})^2+(\frac{a-b}{p-q})^2 \geq 2$

\begin{algorithm}[h!] 
\caption{Cluster fit in EGA checker}
\label{alg:clusterfit}
\hspace*{\algorithmicindent} \textbf{Input:} Initial Cluster List (ICL), and Effective Gripping Area (EGA) \\
\hspace*{\algorithmicindent} \textbf{Output:} Remained Cluster List(RCL) that can fit in EGA.
\begin{algorithmic}[1]
\STATE $RCL \gets \{\}$
\FOR{i $\gets$ 1 to len(ICL)}
    \STATE $min\_area\_rec \gets$ Min\_Area\_Rec(ICL[i])
    \STATE $fit\_in\_flag \gets$ Rec\_in\_Rec($min\_area\_rec$, EGA)
    \IF{$fit\_in\_flag$}
        \STATE $RCL$.append(ICL[i])
    \ENDIF
\ENDFOR
\STATE \bf{return} $RCL$

\end{algorithmic}
\end{algorithm}

In our problem, the cluster rectangle is the yellow rectangle, while the effective gripper area is the red rectangle. After obtaining their widths and heights, we check if the two sets of widths and heights satisfy either of the conditions. If yes, the cluster is kept in the \textit{cluster list (CL)}. Otherwise, it is eliminated. Figure \ref{fig-rec-in-rec}(B) shows one example of the condition a, while Figure \ref{fig-rec-in-rec}(C) shows one example of the condition b. 

\subsection{Cluster Ranking}
\label{step3} 

\begin{algorithm} 
\caption{Cluster Ranking}
\label{alg:clusterrank}
\hspace*{\algorithmicindent} \textbf{Input:} Clusters List (CL) contains all clusters of a given order \\
\hspace*{\algorithmicindent} \textbf{Output:} Ranked Cluster List (RankCL)
\begin{algorithmic}[1]
\STATE $weight\_list \gets \{\}$
\FOR{i $\gets$ 1 to len(CL)}
    \STATE $current\_weight \gets$ Weight\_Sum(CL[i])
    \STATE $weight\_list$.append($current\_weight$)
\ENDFOR
\STATE $RankCL \gets$ SORT(CL, $weight\_list$)
\STATE \textbf{return} $RankCL$
\end{algorithmic}
\end{algorithm}

To quickly find a suitable cluster among all clusters in CL for picking up $k$ objects, we have developed an algorithm (\textbf{Algorithm} \ref{alg:clusterrank}) that will rank them based on their clique orders and the likelihood of finding a collision-free picking pose. Weight\_Sum subroutine calculates the external connection weight sum for every cluster, and SORT subroutine ranks CL based on the weight, higher weight cluster will be ranked lower since it is prone to be closer to external objects. First, the algorithm ranks all clusters in CL based on their clique orders. Since the goal is to pick up $k$ objects, the clusters with the clique order of $k$ should rank highest, followed by clusters with the clique order of $k+1$, and so on. 
The likelihood of finding a collision-free picking pose can be associated with how isolated the cluster is from other objects in the bin. If a cluster is more isolated than another cluster, it is more likely we can find a picking pose from where the arm can lower the gripper to the bottom of the bin collision-free.

We have considered using an existing clique isolation factor proposed in ITO et al. \cite{isolation_factor}. It counts the number of external connections to one clique and then divides it by the clique order. We found it does not fit our problem well since it doesn't consider how close the connected nodes are, which is an important factor in gauging the risk of collision. Therefore, we define a \textit{external crowd index}. It puts more weight on the ones that are close to the clique. The weight $w_i$ are calculated based on Equation \ref{eq-isolation}. 

\begin{equation}
\begin{array}{ll}
\Delta l & = \frac{H_d - width}{5} \\ 
w_i & = 5 - round(\frac{d_i - width}{\Delta l}) \\
wif & = \sum w_i
\end{array}
\label{eq-isolation}
\end{equation}
where $H_d$ is the Neighbor Distance Threshold, and $width$ is the effective width of the object. For each edge that is connected to the clique, its length $d_i$ is converted to weight. The weight is between 1-5. It is inversely proportional to the edge length. The total weight of all edges connected to the clique is its external crowd index. Since it is used for ranking, normalizing it is unnecessary. 

\begin{figure}[h]
    \centering
    \includegraphics[width=1.0\linewidth]{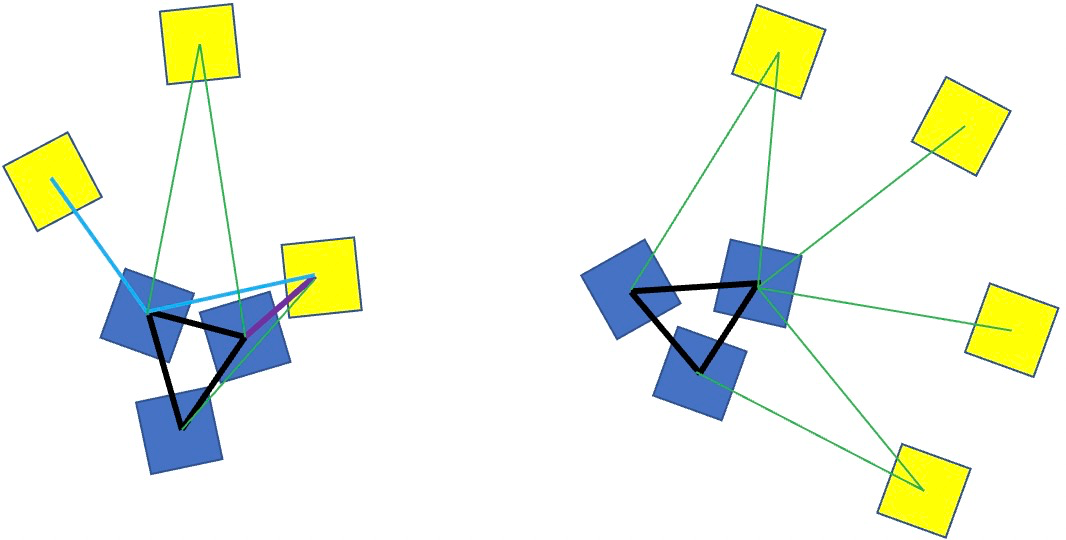}\\
(A)~~~~~~~~~~~~~~~~~~~~~~~~~~~~~~~~~~~(B)
    \caption{Two examples of cluster ranking metrics calculation: Purple line represents edge with weight 3, blue line represents edge with weight 2, green line represents edge with weight 1.}
    \label{ranking_sample}
\end{figure}

Figure \ref{ranking_sample} illustrates how the external crowd index calculation uses the edge distances. Both examples Figure \ref{ranking_sample} have two 3-clusters; their objects are marked as blue, and the neighbor objects to the cluster are marked as yellow. 
The cluster in Figure \ref{ranking_sample} (A) has four edges connecting to three external objects. Based on their lengths, their weights are $5$, $4$, $4$, and $1$. So, its external crowd index is $14$. 
The cluster in Figure \ref{ranking_sample} (B) has six edges connecting to four external objects. Based on their lengths, their weights are all $1$'s. So, its external crowd index is $6$. 
We can see that even though the cluster in Figure \ref{ranking_sample} (B) has four external neighbors and its isolation indicator is worse than the cluster in Figure \ref{ranking_sample} (A), its external crowd index is less. It is consistent with our intuition that picking the cluster in Figure \ref{ranking_sample} (A) has a higher risk of collision than picking the cluster in Figure \ref{ranking_sample} (B). So we rank the cluster in Figure \ref{ranking_sample} (B) higher than in Figure \ref{ranking_sample} (A).

The cliques in CL are ranked based on their clique orders first and then crowd indices to break the ties. The cluster with $k$ nodes and the smallest crowd factor will rank at the top. 

\subsection{Picking Pose Proposal}
\label{step4} 

Once a candidate cluster is selected based on the ranking, our approach will process the layout of the objects in the cluster and then propose several gripper-picking poses that could pick up $k$ objects. We want the approach to propose several poses instead of just one because many of them may not pass the collision check in the next module. 
Since the objects are identical and lying un-stacked in a box, the picking height is fixed and can be calculated. Our approach will focus on obtaining the picking in-plane position and orientation and define them as x, y, and $\gamma$. We compute the cluster center $(c_x, c_y)$ in the world coordinate system and define $\gamma=0^\circ$ when the gripper's $x_g$ axis is aligned with the world $x_w$ axis. The definition of world and gripper axes are shown in Figure \ref{robot-setup}.

\begin{figure}[h!]
    \centering
    \includegraphics[width=1.0\linewidth]{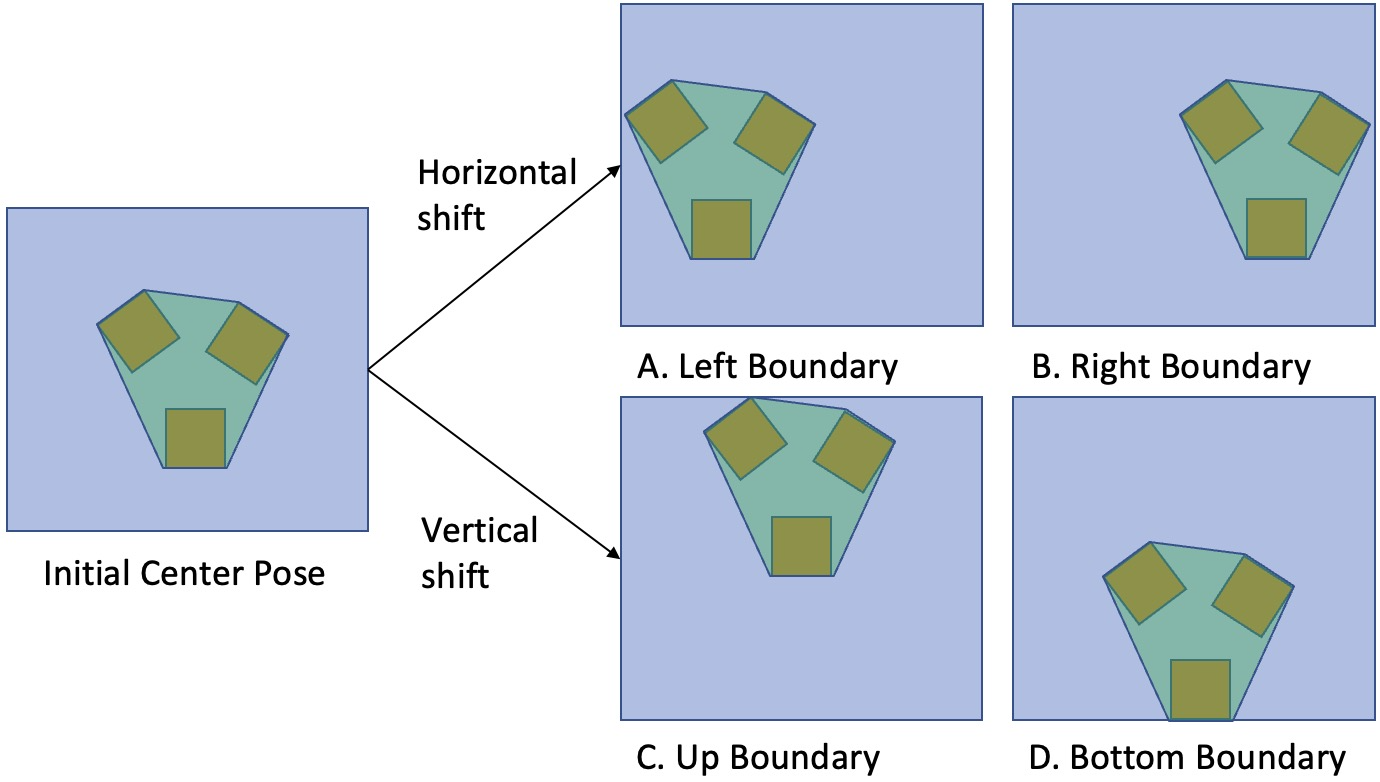}
    \caption{Picking pose sampling procedure: get horizontal and vertical boundaries, then calculate step size and combination of horizontal and vertical shifting to generate picking pose.}
    \label{picking_pose_sampling}
\end{figure}

To propose poses, we first sample 12 $\gamma$'s from $0^\circ$ to $165^\circ$ (the gripper is symmetric) since we found the outcomes of two picking poses are similar if they are different by less than $15^\circ$. 
To sample the positions at each $\gamma_i$, we rotate the gripper by $\gamma$ relative to the world coordinate system and sample along the rotated gripper axes $x_g$ and $y_g$. The sampling range along $x_g$ and $y_g$ are computed to ensure the pose's effective gripping area still encloses the object's convex hull as shown in Figure \ref{picking_pose_sampling}. We found overly fine sampling will increase computation costs and generate many poses that produce the same outcome. Therefore, we sample 10 steps each from the center to the left/right/up/down to their ranges if the ranges are larger than 20 mm. Otherwise, we sample 2 mm for each step since poses with less than a 2 mm difference would produce very similar outcomes. 

Algorithm \ref{alg:posesample} describes the procedure to sample picking poses on a given cluster for 12 degrees. For each rotation, we first get four boundaries of the EGA to cover the cluster using GetBound subroutine. Then, we check if current direction's bound is possible to cover the cluster. If one direction can cover the cluster,  we use GetStepSize subroutine to calculate x and y direction step size according to procedure illustrated above. The sampled picking poses of x,y,$\gamma$ are attached in a list.

\begin{algorithm} 
\caption{Picking Pose Sampling Algorithm}
\label{alg:posesample}
\hspace*{\algorithmicindent} \textbf{Input:} Convexhull Points Set (CPS) of a given cluster, x direction length of EGA in mm (a), y direction length of EGA in mm (b)
 \\
\hspace*{\algorithmicindent} \textbf{Output:} Sampled Picking Poses List (SPL). 
\begin{algorithmic}[1]
\STATE $SPL \gets \{\}$
\FOR{i $\gets$ 0 to 11}
    \STATE $\gamma \gets$ i$\times$15
    \STATE bound\_list $\gets$ GetBound(CPS, $\gamma$)
    \COMMENT{0-3 in bound\_list are left, right, bottom, and up bound}
    \STATE x\_length $\gets$ bound\_list[1]-bound\_list[0]
    \STATE y\_length $\gets$ bound\_list[3]-bound\_list[2]
    \IF{x\_length $>$ a or y\_length $>$ b}
        \STATE \textbf{continue} \COMMENT{at this angle cluster cannot be covered by EGA}
    \ENDIF
    \STATE x\_step, y\_step $\gets$ GetStepSize(x\_length, y\_length)
    \STATE $sampled\_poses \gets \{\}$
    \STATE x$\gets$bound\_list[0]
    \WHILE{x$<$bound\_list[1]}
        \STATE y$\gets$bound\_list[2]
        \WHILE{y$<$bound\_list[3]}
            \STATE $sampled\_poses$.append(x,y,$\gamma$)
            \STATE y += y\_step
        \ENDWHILE
        \STATE x += x\_step
    \ENDWHILE
 \STATE $SPL$.appends($sampled\_poses$)
\ENDFOR
\STATE \textbf{return} $SPL$

\end{algorithmic}
\end{algorithm}

The samples along $x_g$ and $y_g$ are then converted back to the world coordinate system. With $\gamma$, they are associated with the cluster as its picking-pose proposals. 

\subsection{Picking Pose Collision Checking}
\label{step5} 
\begin{figure}
    \centering
    \includegraphics[width=1.0\linewidth]{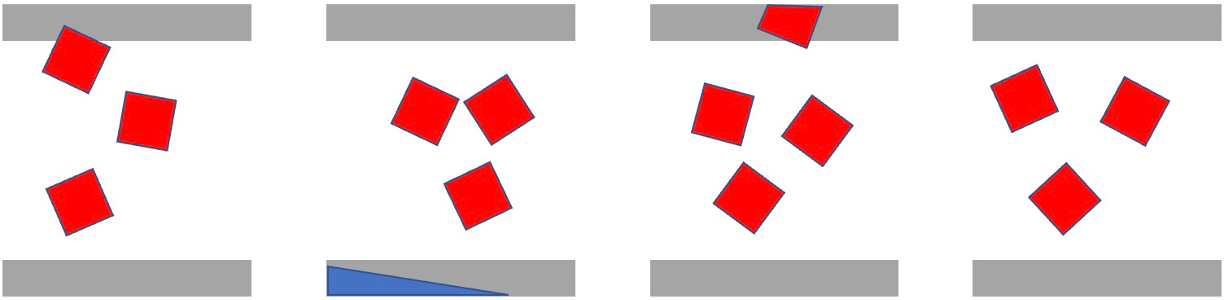}\\
    (A)~~~~~~~~~~~~~~~(B)~~~~~~~~~~~~~~~(C)~~~~~~~~~~~~~~~(D)
    \caption{Collision checking examples: (A) has collision between gripper and internal object, (B) has collision between gripper and picking-bin, (C) has collision between gripper and external object, (D) is collision free.}
    \label{collision_checking}
\end{figure}

We shouldn't select a picking pose that would lead to a collision between the open gripper and the objects. We assume the workspace of the gripper has been confined based on the bin's location and geometry. So, our algorithm checks the collisions in the projected 2D plane since the objects are identical and lying un-stacked. Figure \ref{collision_checking} illustrates three typical collisions and a collision free layout.  (A)-(C) in Figure \ref{collision_checking} are three collision examples with internal object, bin, and external object, (D) is a collision free example. Any proposed picking poses leading to a collision are removed. At this point, we have a cluster list in which each cluster has a list of collision-free picking poses. 

\subsection{Picking-Confidence Estimation}
\label{step6} 

\begin{figure}[h!]
    \centering
    \includegraphics[width=1.0\linewidth]{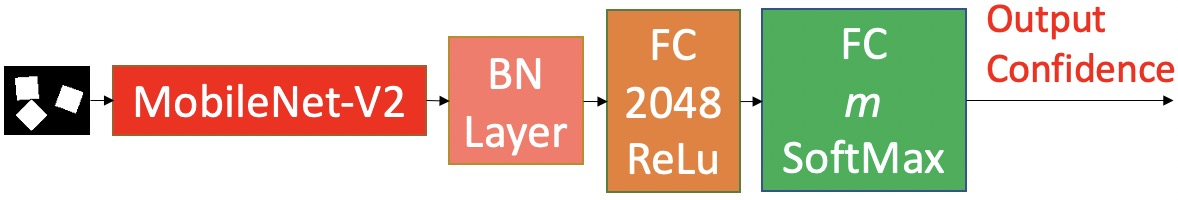}
    \caption{Multi-Object Picking Predictor model architecture.}
    \label{model_archi}
\end{figure}

Using a proposed picking pose, we can obtain a local image of its effective gripping area (defined in Section \ref{step2}). We call it \textit{gripping area image}. We use the pattern in the gripping area image to predict how many objects the gripper will pick up when only picking once. To learn the patterns, we have developed a deep neural network structure that uses the MobileNet-V2\cite{mobilenetv2} for feature extraction and a fully connected (FC) ReLU layer combined with a softmax output layer as a classifier. For efficient training, we add a batch normalization (BN) layer between the MobileNet-V2 and the ReLU layer, as shown in Figure \ref{model_archi}. The output of the network gives the confidences of picking from 0 to $m$ objects. We call the neural network \textit{multi-object picking predictor}.

For each cluster, we input the gripping area images generated using all the proposed picking poses to the multi-object picking predictor in a batch and receive their confidences in picking 0 to $m$ objects. If the desired number of objects is $k$, the picking poses with the highest confidence at the $k$ bit are kept; the rest are discarded. If all proposed poses of a cluster are discarded, the cluster is removed from the CL. 

\paragraph{Data collection and training}
\label{sec-predictor-training}
We train two MOP predictor models: one for the short gripper and one for the long gripper. The short gripper MOP
predictor model is trained with the images of 92,433 random layouts of two small objects: a 1-inch cube and a 2.8-cm cylinder. The long gripper MOP predictor model is trained with the images of 96,836 random layouts of three large objects: a 2-inch cube and 3.8cm cylinder, and a cuboid. The full list of the objects used both in training and testing is in Table \ref{shape_list}. 
Each layout has $2$ - $m$ objects randomly placed in the effective gripping area. Their labels are obtained with their picking outcomes in simulation.
The MobileNet-V2 has been pre-trained on ImageNet. To train the rest of the neural network, we adopt the Adam optimizer with a fixed 1e-4 learning rate and loss function as Categorical Cross-entropy. We set the dropout rate at 0.3 for the fully connected layer. 

\subsection{Selecting pose with high picking confidence}
\label{step7} 

\begin{algorithm}[h!] 
\caption{Action Selection based on confidence}
\label{alg:confidence threshold}
\hspace*{\algorithmicindent} \textbf{Input:} List of clusters to be inspected (CL). Target number $k$. Confidence threshold $H_c$ \\
\hspace*{\algorithmicindent} \textbf{Output:} An action if available, or None if there is no action found.
\begin{algorithmic}[1]
\STATE $backup\_list \gets \{\}$
\FOR{i $\gets$ 1 to len(CL)}
    \STATE sampled\_poses $\gets$ SamplePose(CL[i])
    \STATE valid\_poses $\gets$ CheckCollision(sampled\_poses)
    \STATE local\_images $\gets$ GetLocalImage(valid\_poses)
    \STATE prediction $\gets$ Picking\_Predictor(local\_images)
    \STATE max\_confidence, index $\gets$ Max\_Conf(prediction, $k$)
    \IF{max\_confidence$>H_c$}
        \STATE \textbf{return} valid\_poses[index]
    \ELSIF{max\_confidence$>$0}
        \STATE $backup\_list$.append(index,max\_confidence)
    \ENDIF
\ENDFOR
\IF{$backup\_list!=\{\}$}
    \STATE optimal\_index $\gets$ Max\_Conf\_Index(backup\_list)[0]
    \STATE \textbf{return} valid\_poses[optimal\_index]
\ENDIF
\STATE \textbf{return} None

\end{algorithmic}
\end{algorithm}

So far, our approach processes the object layout in the bin and produces a list (CL) of ranked clusters and their proposed picking pose with their confidences in picking $k$ objects. We could simply go through all clusters in CL and their proposed picking poses and pick the picking pose with the highest confidence of picking $k$ object to execute. However, it is a brute-force approach and could take too much time if many clusters are in a large bin. 

Algorithm \ref{alg:confidence threshold} shows the procedure to find an action based on target number and confidence threshold. SamplePose Subroutine calculates sampled poses based on one cluster. CheckCollision Subroutine filters out poses that have collision with objects or bin. GetLocalImage get the corresponding local image for each picking pose. Max\_Conf Subroutine get the max\_confidence for all local images that predict to pick $k$ objects, index is the corresponding index of this instance . If no pose predicts on $k$, then the max\_confidence will be set as 0. If the max\_confidence for one cluster to pick $k$ objects is above $H_c$, then the picking pose is returned and executed. If the max\_confidence for picking $k$ is below $H_c$, then we store this action as backup until the end if no picking pose's confidence is over the threshold. If all clusters are checked and no action higher than threshold is found to pick $k$ objects, then the action with highest confidence in backup list is selected and passed to robot.
Since efficiency is the goal, we set a good-enough confidence threshold $H_c$ and start checking from the highest-ranked cluster in CL. The selection of the good-enough confidence threshold $H_c$ is described in Section \ref{sec-hc}. When we find a collision-free picking pose with confidence of picking $k$ object over $H_c$, we send it to the robot to execute. This way, we don't need to run through the modules from Picking Pose Proposal to Picking-Confidence Estimation for all clusters in CL. However, for some difficult layouts, it is possible to run through all clusters in CL without seeing one pose with over $H_c$ confidence. Then the algorithm falls back to the brute force approach. 

If the CL is empty because either no cluster has been found or all cluster has been eliminated in the process, the process will report failure and reject the request of picking $k$ objects with one pick. 

\subsection{Parameter Calculation and Selection}
\subsubsection{Neighbor Distance Threshold $H_d$}
\label{edge_threshold_definition}
\begin{figure}[h!]
    \centering
    \includegraphics[width=0.7\linewidth]{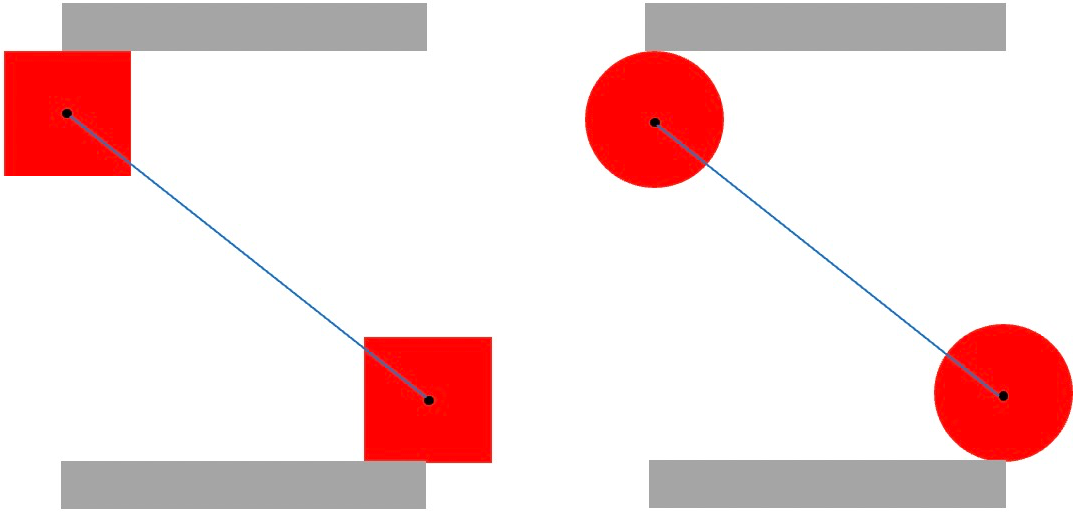}\\
    (A)~~~~~~~~~~~~~~~~~~~~~~(B)
    \caption{Examples of edge threshold selection on cube and cylinder objects.}
    \label{edge_threshold_short}
\end{figure}

If two objects are too far apart, they cannot be picked together by the gripper if only picking once. The farthest distance between two objects that could still be picked together is defined as the Neighbor Distance Threshold $H_d$. When the centers of two objects are at the diagonal corner of the open gripper, they are farthest apart but still could be picked up together, as shown in Figure \ref{edge_threshold_short}. The threshold $H_d$ is computed based on the gripper and object sizes. Figure \ref{edge_threshold_short} (A) and Figure \ref{edge_threshold_short} (B) respectively has 9.5cm and 9.4cm as their threshold.

\subsubsection{Good-Enough Confidence Threshold $H_c$}
\label{sec-hc}

The good-enough confidence threshold $H_c$ is introduced to early stop the search for the cluster and picking pose with the highest confidence of picking $k$ objects. It should be set to limit the computation set without sacrificing much of the success rate. If we set $H_c$ too low, the search will stop very early and a picking pose with a low confidence will be used and the execution outcome would have a lower chance of picking up $k$ objects. On the other hand, if we set $H_c$ too high, the search will skip through many good picking poses and would not stop until going through all clusters and proposed poses. 

\begin{figure}[h!]
    \centering
    \includegraphics[width=1.0\linewidth]{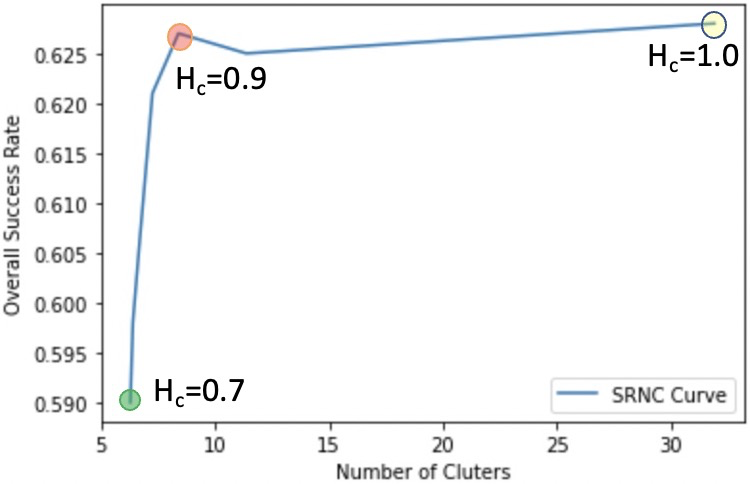}
    \caption{SRNC Curve for picking three 1inch cube, three different $H_c$ are shown, $H_c=0.9$ is selected as the best $H_c$ to achieve optimal Overall Success Rate while sacrificing relatively little in Number of Clusters.}
    \label{srnc_curve}
\end{figure}

To select the proper $H_c$, we designed an experiment to obtain the success-rate and number-of-clusters (SRNC)curve for picking three 1-inch cubes in simulation. The curve is plot with seven thresholds: 0.7, 0.75, 0.8, 0.85, 0.9, 0.95, and 1 (no early stopping). As shown in Figure \ref{srnc_curve}, when the threshold increases, the overall success rate improves while the computation cost also increases. We select $0.9$ as our threshold because its corresponding success rate is close to the one without early stopping, while the averaged computation time is more than 70\% less than without early stopping. The 0.9 threshold is tested on cylinder shape and other different targets in simulation and achieve similar result to balance well between success rate and number of clusters needed to inspect.

We have also plotted the SRNC curves for picking two, three, and four objects. Setting $H_c$ to 0.9 works for all those targets. Therefore, we set $H_c$ to 0.9 for all objects and targets.

\section{Experiments and Evaluation}
\label{Experiments and Evaluation}
\subsection{Setup}
\label{setup_section}
\begin{figure}[h!]
    \centering
    \includegraphics[width=1.0\linewidth]{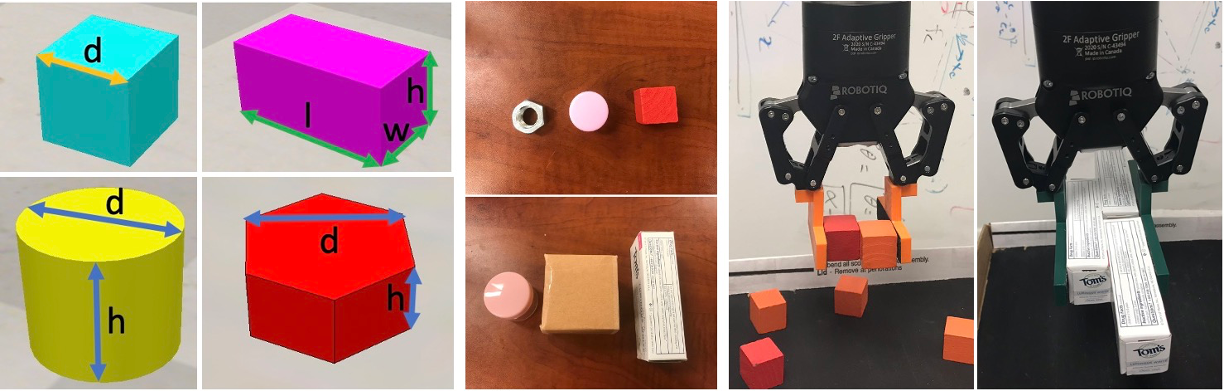}\\
    (A)~~~~~~~~~~~~~~~~~~~~~(B)~~~~~~~~~~~~~~~~~(C)~~~
    \caption{Four types of shapes being used in simulation and their corresponding size metrics. Middle part shows six objects in real testing, the top three objects are used for the short gripper, and the bottom three objects are used for long gripper. Short and Long gripper figures are shown in the right figure.}
    \label{obj_gripper_setup}
\end{figure}

Figure \ref{obj_gripper_setup} shows the setting of our target objects and the grippers. Figure \ref{obj_gripper_setup} (A) shows four types of objects used for evaluation and their corresponding size metrics. Figure \ref{obj_gripper_setup} (B) shows the target objects used to evaluate our algorithms in the real-world setup. We divide all objects into two sets based on their sizes and design grippers for them accordingly. For the three objects in the upper part of Figure \ref{obj_gripper_setup} (B), we design a gripper with $7.5cm$ length and $8.4cm$ spread. The length is around three times as the 1-inch wood cube side length, and this gripper is shown in the left half of Figure \ref{obj_gripper_setup} (C). For the three objects in the lower part in Figure \ref{obj_gripper_setup} (B), we design a gripper with $15.0cm$ length and $8.4cm$ spread. The length is around three times the 2-inch gift box side length. This paper does not focus on designing grippers, but our procedure could be transferred to other objects and different size grippers. 

The base size of the bin used in the real setup is $30.5cm\times38.1cm$. It is the size of a standard storage bin. We select the bin with a short height for better visualization. In the simulation, we use a $38.0cm\times38.0cm$ bin. The bin height is arbitrary as there is collision checking with the container boundary. The two setups have different bin sizes, and our algorithms work equally well on both setups. Our algorithms are not sensitive to the bin size. 

\begin{figure*}[t!]
    \centering
    \includegraphics[width=0.16\linewidth]{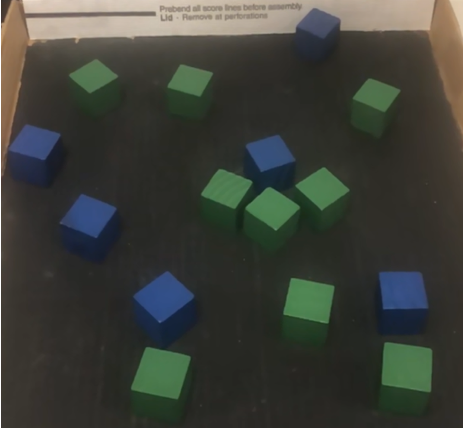}
    \includegraphics[width=0.16\linewidth]{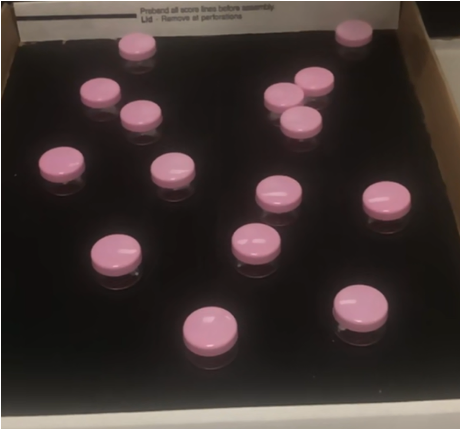}
    \includegraphics[width=0.16\linewidth]{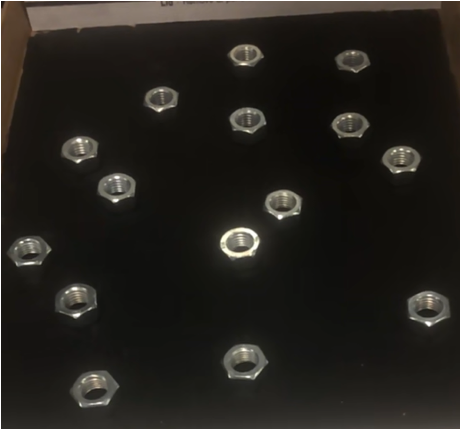}
    \includegraphics[width=0.16\linewidth]{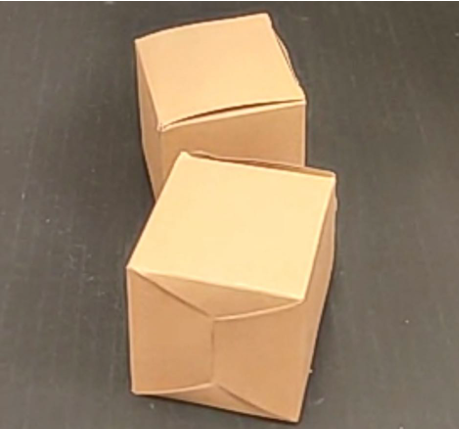}
    \includegraphics[width=0.16\linewidth]{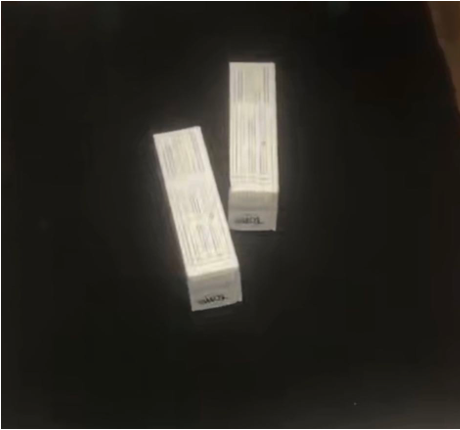}
    \includegraphics[width=0.16\linewidth]{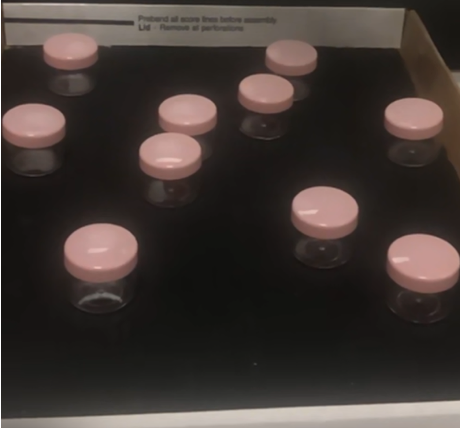}\\
    \includegraphics[width=0.16\linewidth]{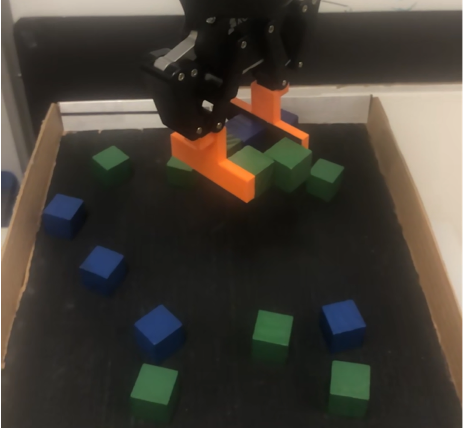}
    \includegraphics[width=0.16\linewidth]{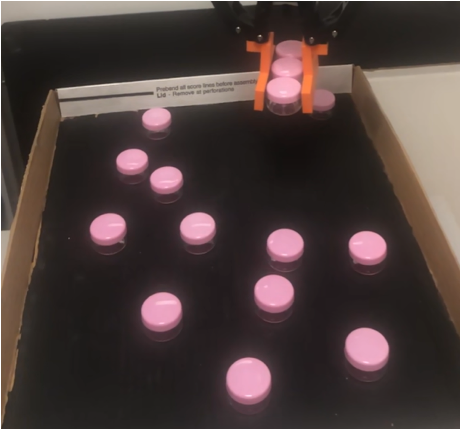}
    \includegraphics[width=0.16\linewidth]{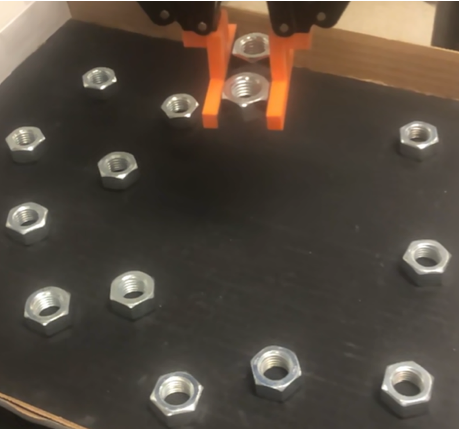}
    \includegraphics[width=0.16\linewidth]{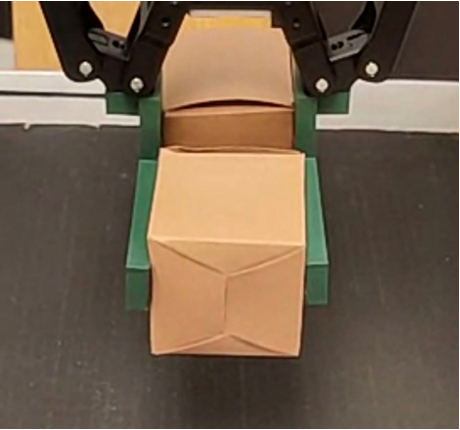}
    \includegraphics[width=0.16\linewidth]{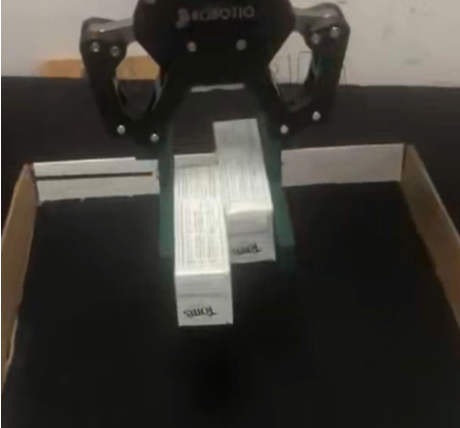}
    \includegraphics[width=0.16\linewidth]{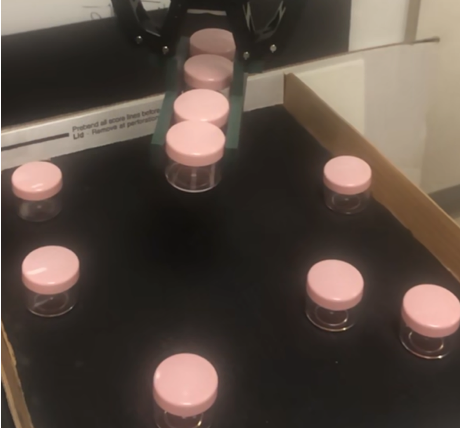}
    \caption{Picking examples in real setup, from left to right: 1inch cube, 2.8cm cylinder, 2.3cm hexagonal nuts, 2inch cube, cuboid, and 3.8cm cylinder}
    \label{real_grasp_examples}
\end{figure*}

Since one important target of the OPOS is logistic applications, we select three common shapes in warehouses. They are cube, cylinder, and cuboid. They are common packaging or container shapes. So, in the real setup, we use cardboard packaging boxes, small- and median-size cosmetic jars, toothpaste boxes, and wood cubes. We then select hexagonal nuts commonly used in manufacturing to evaluate the generalization capability of the proposed approach. The sizes of the objects in the simulation are designed to match the ones in real.
Table \ref{shape_list} lists all shape types, their indices, and the gripper used for picking.
Their dimension parameters are defined in Figure \ref{obj_gripper_setup} (A) and the dimension specifications are shown in Table \ref{shape_list}. The bold font in Table \ref{shape_list} indicates the object is not in the training set, and $*$ indicates the object is also tested in the real setup. Overall, we have evaluated the proposed approach on 12 different objects in simulation and six objects in real. Among them, seven of them are not in the training set. All hexagon objects (nuts) are not in the training set.
Figure \ref{real_grasp_examples} shows examples of multi-object grasping of 6 different objects in real-setup.

\begin{table}
\centering
\caption{Object list table. The meaning of size for each object can be checked from Figure \ref{obj_gripper_setup} (A). We hereafter refer each object by its index. A bold font index means the object is unseen in training stage, and ending * in an index means this object is also used in real testing.}
\label{shape_list}
\begin{tabular}{|c|l|l|c|c|} 
\hline
Shape                     & \multicolumn{1}{c|}{Size(cm)}                                   & \multicolumn{1}{c|}{Index} & Max Num & \multicolumn{1}{l|}{Gripper}  \\ 
\hline
\multirow{4}{*}{Cube}     & d: 2.0                                                          & \textbf{cube\_s\_s}        & 4              & \multirow{3}{*}{Short}        \\ 
\cline{2-4}
                          & d: 2.54                                                         & cube\_m\_s*                & 4              &                               \\ 
\cline{2-4}
                          & d: 3.0                                                          & \textbf{cube\_l\_s}        & 4              &                               \\ 
\cline{2-5}
                          & d: 5.1                                                          & cube\_l*                   & 3              & Long                          \\ 
\hline
\multirow{4}{*}{Cylinder} & \begin{tabular}[c]{@{}l@{}}d: 2.3\\h: 2.5\end{tabular}          & \textbf{cylin\_s\_s}       & 3              & \multirow{3}{*}{Short}        \\ 
\cline{2-4}
                          & \begin{tabular}[c]{@{}l@{}}d: 2.8\\h: 2.5\end{tabular}          & cylin\_m\_s*               & 3              &                               \\ 
\cline{2-4}
                          & \begin{tabular}[c]{@{}l@{}}d: 3.3\\h: 2.5\end{tabular}          & \textbf{cylin\_l\_s}       & 3              &                               \\ 
\cline{2-5}
                          & \begin{tabular}[c]{@{}l@{}}d: 3.8\\h: 3.0\end{tabular}          & cylin\_l*                  & 4              & Long                          \\ 
\hline
Cuboid                    & \begin{tabular}[c]{@{}l@{}}l: 10.6\\w: 3.0\\h: 3.5\end{tabular} & cuboid\_l*                 & 4              & Long                          \\ 
\hline
\multirow{3}{*}{Hexagon}  & \begin{tabular}[c]{@{}l@{}}d: 2.0\\h: 1.0\end{tabular}          & \textbf{hexa\_s\_s}        & 4              & \multirow{3}{*}{Short}        \\ 
\cline{2-4}
                          & \begin{tabular}[c]{@{}l@{}}d: 2.3\\h: 1.0\end{tabular}          & \textbf{hexa\_m\_s*}       & 4              &                               \\ 
\cline{2-4}
                          & \begin{tabular}[c]{@{}l@{}}d: 2.3\\h: 1.0\end{tabular}          & \textbf{hexa\_l\_s}        & 4              &                               \\
\hline
\end{tabular}
\end{table}

\subsection{Evaluation Metrics and Protocols}
Since the goal is to pick up $k$ objects efficiently and accurately for any arbitrary scenes, we define the following evaluation metrics: 
\begin{enumerate}
    \item Availability rate (AR): among all arbitrary scenes, the percentages of the scenes for which the OPOS believes it could pick up $k$ objects at once;
    \item Execution success rate (ESR): among all available scenes for picking $k$ objects, the percentages of the scenes for which the OPOS actually picks up exact $k$ objects;
    \item Overall success rate (OSR): among all arbitrary scenes, the percentages of the scenes for which the OPOS picks up $k$ objects with one picking motion, OSR is the multiplication of AR and ESR: $OSR = AR\times ESR$;
    \item Number of picking motions (NP): the number of picking motions needed to pick up exact $k$ objects if not required to only pick once.
\end{enumerate}

The difference between OSR and ESR lies at the scenes, for which our approach fails to produce any picking pose. It could happen when the algorithm cannot find a cluster that contains $k$ or more objects. For example, it happens when all objects are sparsely scattered in a bin, and they are all farther apart from each other than the gripper length. Since this paper does not consider pushing motions to gather the objects, our approach would not produce any picking pose for this example. It could also happen when the algorithm cannot find a collision-free picking pose when there are too many objects in the bin, and the gripper is large. 

The number of picking motions (NP) provides a direct measure of the efficiency of the approach. The NP of a traditional single-object picking (SOP) approach would be equal to or above $k$. The proposed MOP approach could pick up $k$ objects with one picking motion for some scenes but may need several picking motions to recover from failures so that the total number of the picked objects is exactly $k$. If the proposed MOP has an averaged NP lower than $k$, it would be considered more efficient than the SOP.

To rigorously evaluate the proposed approaches, we design three evaluation protocols. In the simulation, to obtain reliable OS rates and ES rates in different situations, we follow this protocol:
\begin{itemize}
    \item Train MOP predictors. We train two MOP predictors as described in Section \ref{sec-predictor-training}. 
    \item Create random layouts. For a small object in testing, we randomly create 1,000 layouts in five densities - having 20, 25, 30, 35, and 40 objects in the bin. For the median object (3.8 cm cylinder) in testing, we also randomly create 1000 layouts in three densities - with 10, 15, 20, 25, 30 objects in the bin. We use the testing layouts collected when we train the long-gripper MOP predictor for large objects. 
    \item Run MOP algorithms. We run the proposed MOP algorithms on the test layouts given a desired number of objects. The number ranges from $2$ to the max number indicated in the column ``Max Num'' in table \ref{shape_list}. The max numbers are selected based on how many objects the gripper can pick up in reality. 
    \item Collect evaluation data. For each run, we observe and record: the desired number of objects (target), the object index, the number of objects in the bin, if the algorithms find a picking pose, and how many are actually picked up. 
\end{itemize}

In the real environment, we directly use the MOP predictors trained in simulation without transfer learning in real-setup. So its protocol doesn't have the "Train MOP predictors" module. The layout numbers are 80 for each object and each density. The rest is the same as the one in the simulation. 

To evaluate the capability of generalization, we designed a protocol to test hexagons in both simulation and real. We use the short gripper MOP predictor trained in simulation with other shapes. The rest of the protocols are the same as the ones in the simulation and real.

\subsection{Success Rate Results}
\label{core_evaluation}

\begin{table}[h!]
\centering
\caption{Simulation Success Rate on Trained Objects.}
\label{sim_core_result}
\begin{tabular}{|c|c|c|c|c|c|} 
\hline
\multirow{2}{*}{\begin{tabular}[c]{@{}c@{}}Setting\\(num of scene)\end{tabular}} & \multirow{2}{*}{\begin{tabular}[c]{@{}c@{}}Obj Index\\(density)\end{tabular}} & \multirow{2}{*}{Result} & \multicolumn{3}{c|}{Target}               \\ 
\cline{4-6}
                                                                                 &                                                                               &                         & 2        & 3       & 4                    \\ 
\hline
\multirow{9}{*}{\begin{tabular}[c]{@{}c@{}}Simulation\\(200)\end{tabular}}       & \multirow{3}{*}{\begin{tabular}[c]{@{}c@{}}cube\_m\_s\\(20)\end{tabular}}     & AR                      & 100.00\% & 77.00\% & 10.50\%              \\ 
\cline{3-6}
                                                                                 &                                                                               & ESR                     & 97.50\%  & 74.03\% & 61.90\%              \\ 
\cline{3-6}
                                                                                 &                                                                               & OSR                     & 97.50\%  & 57.00\% & 6.50\%               \\ 
\cline{2-6}
                                                                                 & \multirow{3}{*}{\begin{tabular}[c]{@{}c@{}}cylin\_m\_s\\(20)\end{tabular}}    & AR                      & 100.00\% & 95.00\% & \multirow{3}{*}{NA}  \\ 
\cline{3-5}
                                                                                 &                                                                               & ESR                     & 96.50\%  & 94.21\% &                      \\ 
\cline{3-5}
                                                                                 &                                                                               & OSR                     & 96.50\%  & 89.50\% &                      \\ 
\cline{2-6}
                                                                                 & \multirow{3}{*}{\begin{tabular}[c]{@{}c@{}}cylin\_l\\(10)\end{tabular}}       & AR                      & 100.0\%  & 79.50\% & 8.50\%               \\ 
\cline{3-6}
                                                                                 &                                                                               & ESR                     & 97.50\%  & 96.86\% & 94.12\%              \\ 
\cline{3-6}
                                                                                 &                                                                               & OSR                     & 97.50\%  & 77.00\% & 8.00\%               \\
\hline
\end{tabular}
\end{table}

\begin{table}[h!]
\centering
\caption{Real World Success rate on Trained Objects.}
\label{real_core_result}
\begin{tabular}{|c|c|c|c|c|c|} 
\hline
\multirow{2}{*}{\begin{tabular}[c]{@{}c@{}}Setting\\(num of scene)\end{tabular}} & \multirow{2}{*}{\begin{tabular}[c]{@{}c@{}}Obj Index\\(density)\end{tabular}} & \multirow{2}{*}{Result} & \multicolumn{3}{c|}{Target}              \\ 
\cline{4-6}
                                                                                 &                                                                               &                         & 2       & 3       & 4                    \\ 
\hline
\multirow{9}{*}{\begin{tabular}[c]{@{}c@{}}Real\\(80)\end{tabular}}              & \multirow{3}{*}{\begin{tabular}[c]{@{}c@{}}cube\_m\_s\\(15)\end{tabular}}     & AR                      & 97.50\% & 66.25\% & 16.25\%              \\ 
\cline{3-6}
                                                                                 &                                                                               & ESR                     & 96.15\% & 77.36\% & 76.92\%              \\ 
\cline{3-6}
                                                                                 &                                                                               & OSR                     & 93.75\% & 51.25\% & 12.50\%              \\ 
\cline{2-6}
                                                                                 & \multirow{3}{*}{\begin{tabular}[c]{@{}c@{}}cylin\_m\_s\\(15)\end{tabular}}    & AR                      & 97.50\% & 90.00\% & \multirow{3}{*}{NA}  \\ 
\cline{3-5}
                                                                                 &                                                                               & ESR                     & 94.87\% & 95.83\% &                      \\ 
\cline{3-5}
                                                                                 &                                                                               & OSR                     & 92.50\% & 86.25\% &                      \\ 
\cline{2-6}
                                                                                 & \multirow{3}{*}{\begin{tabular}[c]{@{}c@{}}cylin\_l\\(10)\end{tabular}}       & AR                      & 83.75\% & 43.75\% & 11.25\%              \\ 
\cline{3-6}
                                                                                 &                                                                               & ESR                     & 97.01\% & 91.43\% & 88.89\%              \\ 
\cline{3-6}
                                                                                 &                                                                               & OSR                     & 81.25\% & 40.00\% & 10.00\%              \\
\hline
\end{tabular}
\end{table}

Table \ref{sim_core_result} and \ref{real_core_result} show the result of the AR, ESR, and OSR in simulation and the real world, respectively. 
The number in the Obj Index column is density of objects in the bin before picking. Both tables show the result on three objects seen in training stage: $cube\_m\_s$, $cylin\_m\_s$, $cylin\_l$.

Table \ref{sim_core_result} shows the result of AR, ESR, and OSR for three objects above in simulation. 

It shows that OSR and ESR for all three objects are higher or equal to 96.50\% when target\#=2. When target\#=3, OSR and ESR for $cube\_m\_s$, $cylin\_m\_s$, $cylin\_l$ are 57.00\% and 74.03\%, 89.50\% and 94.31\%, 77.00\% and 96.86\%. When target\#=4, OSR and ESR for $cube\_m\_s$, $cylin\_l$ are 6.50\% and 61.90\%, 8.00\% and 94.12\%, the result for $cylin\_m\_s$ is $NA$ as the Max Num for this object is 3. The AR value of $cube\_m\_s$ for target\#=2, 3, and 4 are 100.00\%, 77.00\% and 10.50\%. A clear trend is that for each setting(identical object, identical density), AR value significantly decreases when the target number $k$ increases.

In Real Result Table \ref{real_core_result}, the result shows that OSR and ESR for object $cube\_m\_s$, $cylin\_m\_s$, $cylin\_l$ are 93.75\% and 96.15\%, 92.50\% and 94.87\%, 81.25\% and 97.01\% when target\#=2. When target\#=3, OSR and ESR for $cube\_m\_s$, $cylin\_m\_s$, $cylin\_l$ are 51.25\% and 77.36\%, 86.25\% and 95.83\%, 40.00\% and 91.43\%. When target\#=4, OSR and ESR for $cube\_m\_s$, $cylin\_l$ are 12.50\% and 76.92\%, 10.00\% and 88.89\%, the result for $cylin\_m\_s$ is $NA$ as the Max Num for this object is 3. The pattern of AR for different target number is identical to simulation result.

The result from the two tables indicates the followings:
\begin{itemize}
    \item OSR and ESR are high for all objects under simulation and real setup when target\#=2.
    \item When target\#=3 or 4, both OSR and ESR decrease for all objects, and the gap between these two values becomes larger because of AR value decreasing when target number increasing, and the reason AR decreases is that there are more unavailable cases without a solution.
    \item ESR for larger target number decreases for the same setting because of randomness and the picking number estimator accuracy.
    \item Under the same setup (identical object, identical density), it is harder to find an available cluster of a larger target number which is one reason causing low AR and thereafter the bigger difference between OSR and ESR. For an instance, The OSR rate for $cube\_m\_s$ in Table \ref{sim_core_result} is 97.50\%, 57.00\%, and 6.50\% for target\#=2, 3, and 4 under same density 20. The ESR rate for the three densities are 97.50\%, 74.03\%, and 61.90\%. The decrement in OSR is due to one reason there are fewer 3 and 4 clusters under 20 objects density.
    \item Another reason causing lower OSR for larger target numbers is more collision when trying to pick a higher number. In following contents, we analyze how density affects OSR in Section
    \ref{diff_density_section}.
    \item Furthermore, for each setting (same object, same density), the ESR decreases when target number increases, this is another reason causing OSR decreasing. The reason ESR decreases as target number increases is because there are more randomness when picking higher number of objects, therefore the predictor is less accurate on more objects picking.
\end{itemize}

\begin{figure}[h!]
    \centering
    \includegraphics[width=1\linewidth]{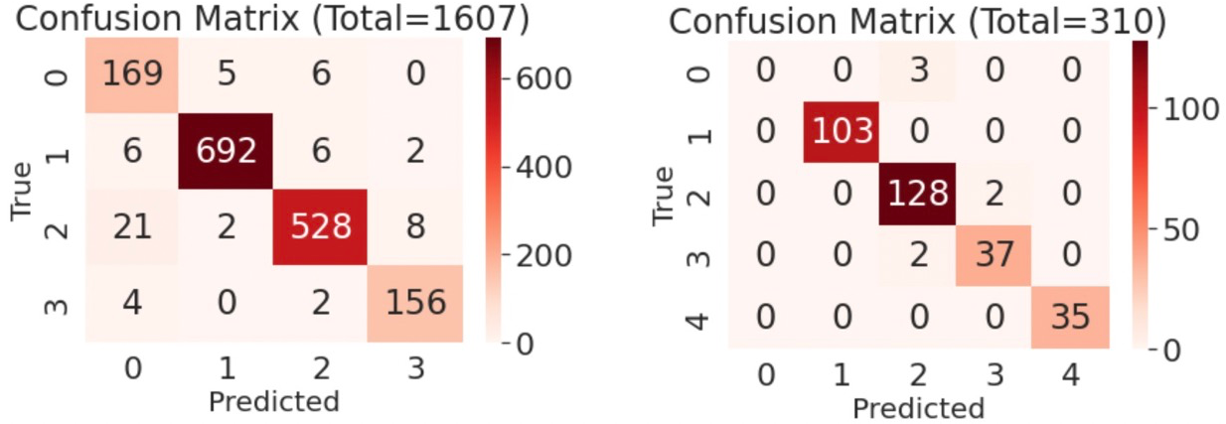}\\
    (A)~~~~~~~~~~~~~~~~~~~~~~~~~~~~~~~~~~~(B)
        \includegraphics[width=1\linewidth]{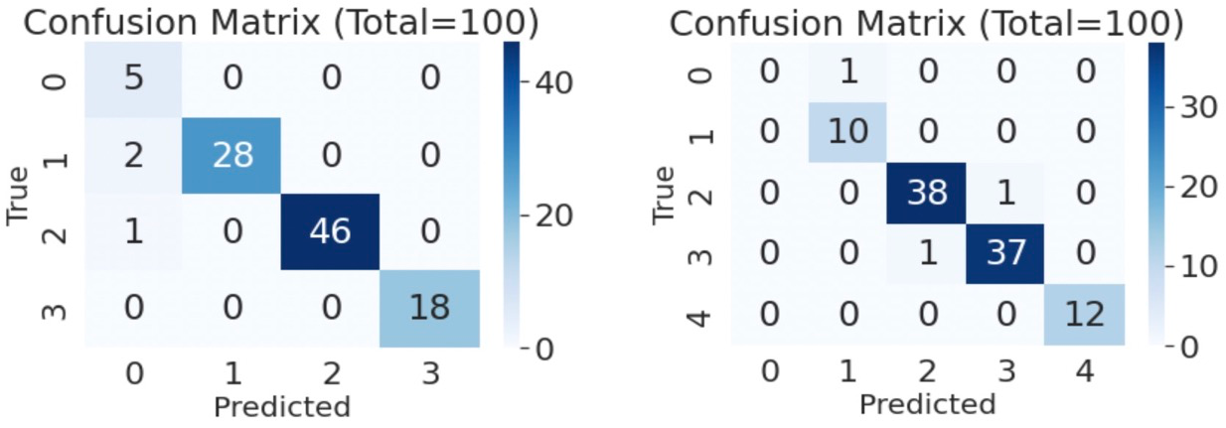}\\
        (C)~~~~~~~~~~~~~~~~~~~~~~~~~~~~~~~~~~~(D)
    \caption{Confusion Matrix for 2inch cube and cuboid in simulation testing and real testing. Top left is 2inch cube in simulation. Top right is cuboid in simulation. Bottom left is cube in real. Bottom right is cuboid in real.}
    \label{confusion_matrix}
\end{figure}

Since it is difficult to fit many large objects: $cube\_l$ and $cuboid\_l$ in the bin loosely, we use test layouts for the longer griper MOP predictor for evaluation. Figures \ref{confusion_matrix} (A) and (B) show the confusion matrix of the testing set in simulation for $cube\_l$ and $cuboid\_l$; Figures \ref{confusion_matrix} (C) and (D) show the confusion matrix in the real setting for $cube\_l$ and $cuboid\_l$. The overall MOP predictor success rates for $cube\_l$ and $cuboid\_l$ are 96.14\% and 97.74\% in simulation, 97.00\% and 97.00\% in real setup. 

\subsection{Efficiency Evaluation}
\label{app-example}

To compute the number of picking motions (NP), we define a hypothetical picking and transferring procedure that handles picking failures to make sure the system picks and transfers exactly $k$ objects. We assume $k$ could be 2, 3, or 4. The procedure runs the proposed approach to searches for a picking pose for $p=k$. If it cannot find a pose, it will search for a picking pose for $p=k-1$ and so on, until $p=1$. If the approach finds a pose, it will be executed, and the output will be either successful or not. If it is successful, the procedure will pick the remaining $k-p$ using single-object picking (SOP). Otherwise, the procedure will handle two kinds of failures in the following ways: 
\begin{itemize}
    \item  Failure type 1 (FT1)- the number of the picked objects $q$ is smaller than $k$ (including nothing is picked), the procedure runs SOP to pick the remaining $k-q$ objects.
    \item Failure type 2 (FT2)- the number of the picked objects $q$ is larger than $k$, the procedure runs SOP to pick up $q-k$ objects from the receiving bin.
\end{itemize}

So, for the cases when $p=k$ and are successful, their number of picking motions is $1$. For the cases having FT1, their number of picking motions is $1+k-q$. For the cases having FT2, their number of picking motions is $1+q-k$. We assume SOP has a 100\% success rate. Using the statistics obtained for OSR and ESR, we can compute the hypothetical averaged number of picking numbers to measure the efficiencies of MOP and SOP. Table \ref{distribution table} shows the result of retrieving $k$ objects for $cube\_m\_s$ and $cylin\_m\_s$ in real world and simulation. We use a consistent density of 15 in real testing for both objects and five densities (from 20 to 40 with a step size of 5) in the simulation to investigate how the density of objects may affect the result.

\begin{table}[h!]
\centering
\caption{Averaged Number of Picking Motion.}
\label{distribution table}
\begin{tabular}{|c|c|c|l|l|l|l|l|} 
\hline
\multicolumn{2}{|l|}{}                                         & \multicolumn{6}{c|}{Density}                                                                                                                \\ 
\hline
Object                       & \multicolumn{1}{l|}{\textit{k}} & 15(real) & \multicolumn{1}{c|}{20} & \multicolumn{1}{c|}{25} & \multicolumn{1}{c|}{30} & \multicolumn{1}{c|}{35} & \multicolumn{1}{c|}{40}  \\ 
\hline
\multirow{3}{*}{cube\_m\_s}  & 2                               & 1.075    & 1.045                   & 1.035                   & 1.065                   & 1.050                   & 1.050                    \\ 
\cline{2-8}
                             & 3                               & 1.650    & 1.585                   & 1.650                   & 1.570                   & 1.475                   & 1.480                    \\ 
\cline{2-8}
                             & 4                               & 2.350    & 2.470                   & 2.510                   & 2.435                   & 2.315                   & 2.145                    \\ 
\hline
\multirow{2}{*}{cylin\_m\_s} & 2                               & 1.088    & 1.040                   & 1.040                   & 1.055                   & 1.035                   & 1.075                    \\ 
\cline{2-8}
                             & 3                               & 1.200    & 1.150                   & 1.085                   & 1.145                   & 1.095                   & 1.115                    \\
\hline
\end{tabular}
\end{table}

Overall, if the task is to pick and transfer two objects, on average, it would take our approach less than 1.1 picking motions to pick and transfer exactly two objects, while the SOP would need $2$. If the task is to pick and transfer three 1-inch cubes, on average, it will take our approach less than $1.7$ picking motions to pick and transfer exactly three objects, while the SOP would need $3$. If the task is to pick and transfer four 1-inch cubes, on average, it will take our approach less than $2.6$ picking motions to pick and transfer exactly four objects, while the SOP would need $4$. For the small cosmetic jar (cylin\_m\_s), when asked to pick and transfer exact three jars, it takes our approach, on average, $1.085$ to $1.150$ picking motions to pick and transfer exactly three jars, while the SOP would need $3$. It shows because of the capacity of the short gripper. The best strategy is to pick either 2 or 3 objects at a time if the desired number $k$ is over $3$.

We can also see a trend -- when the density increases, the average picking trials decreases. It is because the low-density layouts are less likely to provide a decent number of clusters. We further study the effect of the density on the success rates next.

\subsection{Study on Density Effect}
\label{diff_density_section}

\begin{table}[h!]
\centering
\caption{Picking result under different object density.}
\label{diff_density_table}
\begin{tabular}{|c|c|c|c|c|} 
\hline
\multicolumn{2}{|c|}{}       & \multicolumn{3}{c|}{Target Picking Num}  \\ 
\hline
Density             & Result & 2        & 3       & 4                   \\ 
\hline
\multirow{3}{*}{20} & AS     & 100.00\% & 77.00\% & 10.50\%             \\ 
\cline{2-5}
                    & ESR    & 97.50\%  & 74.03\% & 61.90\%             \\ 
\cline{2-5}
                    & OSR    & 97.50\%  & 57.00\% & 6.50\%              \\ 
\hline
\multirow{3}{*}{25} & AS     & 100.00\% & 84.00\% & 14.50\%             \\ 
\cline{2-5}
                    & ESR    & 97.00\%  & 67.86\% & 68.97\%             \\ 
\cline{2-5}
                    & OSR    & 97.00\%  & 57.00\% & 10.00\%             \\ 
\hline
\multirow{3}{*}{30} & AS     & 100.00\% & 90.50\% & 21.50\%             \\ 
\cline{2-5}
                    & ESR    & 96.00\%  & 72.93\% & 58.14\%             \\ 
\cline{2-5}
                    & OSR    & 96.00\%  & 66.00\% & 12.50\%             \\ 
\hline
\multirow{3}{*}{35} & AR     & 100.00\% & 91.00\% & 30.50\%             \\ 
\cline{2-5}
                    & ESR    & 97.00\%  & 71.98\% & 67.21\%             \\ 
\cline{2-5}
                    & OSR    & 97.00\%  & 65.50\% & 20.50\%             \\ 
\hline
\multirow{3}{*}{40} & AR     & 100.00\% & 90.50\% & 32.50\%             \\ 
\cline{2-5}
                    & ESR    & 96.00\%  & 75.14\% & 76.92\%             \\ 
\cline{2-5}
                    & OSR    & 96.00\%  & 68.00\% & 25.00\%             \\
\hline
\end{tabular}
\end{table}
As shown in previous tables, the number of objects inside the picking bin may greatly affect the picking result and accuracy, especially when the target number is higher. Therefore, we have conducted simulation experiments to explore how different densities affect picking results. The study has been done on all 12 objects. The density effects are similar for all of them. So, here we show the results on $cube\_m\_s$ as an example in Table \ref{diff_density_table}. 

When the target picking number is 2, there is no significant difference in OSR or ESR among different object densities, this is because AR very high since finding a pose to pick 2 objects is easy. The lowest OSR and ESR are both 96.00\%, and the highest OSR and ESR are both 97.50\%.
When the target number is 3 or 4, we can see the AR strictly increases when the density increases, this causes OSR increases as well since ESR does not have a clear trend of changing for different density. This is because when more objects are in the bin, there are more candidate clusters to select from.

Overall, since high density provides more clusters, the proposed approach is more likely to find a suitable cluster, especially for picking 3 and 4 objects. It doesn't affect picking two because clusters of two are common and a picking pose is easier to find compared to higher number picking.
The ESR rate is not associated with the density because our MOP predictors have a consistent accuracy over different densities.

\subsection{Ablation Study}
\label{baseline}
\begin{table*}[t!]
\centering
\caption{Ablation Study Table.}
\label{ablation_table}
\begin{tabular}{|c|l|l|l|l|l|l|} 
\hline
\multicolumn{1}{|l|}{\multirow{2}{*}{}} & \multicolumn{6}{c|}{Density}                                                                                                                                   \\ 
\cline{2-7}
\multicolumn{1}{|l|}{}                  & \multicolumn{1}{c|}{20} & \multicolumn{1}{c|}{25} & \multicolumn{1}{c|}{30} & \multicolumn{1}{c|}{35} & \multicolumn{1}{c|}{40} & \multicolumn{1}{c|}{Average}  \\ 
\hline
Setting                                 & \multicolumn{6}{c|}{Result (num of inspected clusters/OSR/ESR)}                                                                                                   \\ 
\hline
\textbf{*PA}                            & 2.76/57.00\%/74.03\%    & 4.67/57.00\%/67.86\%    & 7.15/66.00\%/72.93\%    & 11.69/65.50\%/71.98\%   & 15.71/68.00\%/71.58\%   & 8.40/62.70\%/71.68\%          \\ 
\hline
B-1                                     & 4.02/56.00\%/72.73\%    & 7.85/57.00\%/67.86\%    & 13.24/64.50\%/71.27\%   & 19.16/66.00\%/72.53\%   & 28.08/68.00\%/71.58\%   & 14.47/62.30\%/71.19\%         \\ 
\hline
B-2                                     & 2.30/51.50\%/66.88\%    & 2.72/49.50\%/58.93\%    & 3.95/61.00\%/67.40\%    & 4.37/61.50\%/67.58\%    & 5.73/59.50\%/62.63\%    & 3.81/56.60\%/64.68\%          \\ 
\hline
B-3                                     & 3.21/51.00\%/66.23\%    & 7.33/48.50\%/57.74\%    & 10.07/61.00\%/67.40\%   & 16.92/61.00\%/67.03\%   & 23.37/60.00\%/63.16\%   & 12.18/56.30\%/64.31\%         \\
\hline
\end{tabular}
\end{table*}

The proposed approach (PA) could be simplified by eliminating the cluster ranking module and confidence threshold in searching for the picking pose. To demonstrate the benefit of including them, we define three baseline approaches. The baseline \#1 (B-1) does not contain the cluster ranking module but uses the good-enough confidence threshold for early stopping. The baseline \#2 (B-2) has the cluster ranking module, but set no confidence threshold, it returns the first found action that is predicted to pick $k$ according to the MOP predictor. The baseline \#3 (B-3) does not contain the cluster ranking module and exhaustively searches through all clusters.

We performed three ablation studies in simulation on $cube\_m\_s$ in five different densities with target picking number 3. The study results are shown in Table \ref{ablation_table}. They are separated by $/$. The first value represents the number of searched clusters before finding the action. The second value is OSR. The third value is ESR. All numbers are the average numbers of 200 random scenes for a given density. Different algorithms are tested on the same series of random scenes. 
The table shows five different density results and the average number of five different density results. 

When the density is 20, \textbf{PA} takes 2.76 clusters on average to find an action to pick three cubes, which is 31.4\% less than B-1 and 20.0\% more than B-2. As for picking success rate, ESR for \textbf{PA} is 57.00\% for OSR and 74.03\% for ESR, 5.50\% and 7.15\% higher compared to B-1, and 1.00\% and 1.30\% higher compared to B-1. B-3 performs the worst for both cluster searching efficiency and success rate. As the object density increases, the difference between \textbf{PA} and algorithms without ranking becomes larger since the layout becomes much more complicated. 

On average, for all five densities, \textbf{PA} reduces 41.95\% number of searched clusters compared to B-1, and 6.10\% and 7.00\% higher accuracy on OSR and ESR compared to B-2. To make a conclusion for the ablation study, \textbf{PA} performs the best in success rate with sacrificing a bit more computation time compared to B-2, however as our algorithm processes each cluster pretty fast, each action generation for \textbf{PA} usually takes less than 3-4 seconds and is tolerable. The cluster ranking module is a key part of the approach to increase searching efficiency without sacrificing anything. 

Our algorithm on average takes 8.40 clusters to search and achieves a 62.70\% overall accuracy, which is a 41.95\% save compared to no cluster ranking baseline and a 6.10\% improvement compared to no confidence threshold. This shows that our cluster ranking algorithm and confidence threshold play important roles in reducing clusters to search and improving overall success rate.

\subsection{Generalization Study}
\label{extensive_study}
To evaluate if the trained object picking number estimation model can generalize to unseen size and shape objects, we design and test a series of experiments to measure the performance of our algorithm on different size and shape objects. 

\subsubsection{Unseen Sizes}
\label{diff_size_section}

\begin{table}[h!]
\centering
\caption{Generalization study of objects with unseen sizes.}
\label{extensive_diff_size}
\begin{tabular}{|c|c|c|c|c|c|} 
\hline
\multicolumn{3}{|c|}{}                                                                                                                                           & \multicolumn{3}{c|}{Target}              \\ 
\hline
Setting                                                                    & Obj Index                                                                  & Result & 2       & 3       & 4                    \\ 
\hline
\multirow{8}{*}{\begin{tabular}[c]{@{}c@{}}Simulation\\(200)\end{tabular}} & \multirow{2}{*}{\begin{tabular}[c]{@{}c@{}}cube\_s\_s\\(20)\end{tabular}}  & OSR    & 94.50\% & 66.50\% & 8.50\%               \\ 
\cline{3-6}
                                                                           &                                                                            & ESR    & 94.50\% & 74.30\% & 70.83\%              \\ 
\cline{2-6}
                                                                           & \multirow{2}{*}{\begin{tabular}[c]{@{}c@{}}cube\_l\_s\\(20)\end{tabular}}  & OSR    & 97.50\% & 24.00\% & 2.00\%               \\ 
\cline{3-6}
                                                                           &                                                                            & ESR    & 97.99\% & 57.14\% & 66.67\%              \\ 
\cline{2-6}
                                                                           & \multirow{2}{*}{\begin{tabular}[c]{@{}c@{}}cylin\_s\_s\\(20)\end{tabular}} & OSR    & 96.50\% & 87.50\% & \multirow{4}{*}{NA}  \\ 
\cline{3-5}
                                                                           &                                                                            & ESR    & 96.50\% & 90.21\% &                      \\ 
\cline{2-5}
                                                                           & \multirow{2}{*}{\begin{tabular}[c]{@{}c@{}}cylin\_l\_s\\(20)\end{tabular}} & OSR    & 96.50\% & 58.50\% &                      \\ 
\cline{3-5}
                                                                           &                                                                            & ESR    & 96.98\% & 72.22\% &                      \\
\hline
\end{tabular}
\end{table}

Table \ref{extensive_diff_size} shows the simulation result on different size objects unseen during the training stage. $cube\_s\_s$ and $cube\_l\_s$ are the smaller and larger version of original trained $cube\_m\_s$ object. 

The result shows that $cube\_s\_s$ achieves relatively similar results in all three different target number picking experiments: 3.00\% lower and 3.00\% lower than $cube\_m\_s$ in OSR and ESR when the target is 2, 9.50\% higher and 0.27\% higher than $cube\_m\_s$ in OS and ES when the target number is 3, 2.00\% higher and 8.93\% higher than $cube\_m\_s$ in OS and ES when the target number is 4. 

When testing on $cube\_l\_s$, target number 2 result is not worse than $cube\_m\_s$. However, the OSR decreases significantly in target 3 and 4 picking experiments. This is because $cube\_l\_s$ is about 20\% larger than $cube\_m\_s$ and causes picking more objects harder. Meanwhile, larger object occupies more space and leaves fewer available picking actions.

The result for $cylin\_s\_s$ and $cylin\_l\_s$ is similar to the result $cylin\_m\_s$ in target 2 picking. When target number is 3, $cylin\_s\_s$ is 2.00\% lower than $cylin\_m\_s$ in OSR and 4.00\% lower than $cylin\_m\_s$ in ESR. $cylin\_l\_s$ is 31.00\% lower than $cylin\_m\_s$ in OSR and 21.99\% lower than $cylin\_m\_s$ in ESR.

This extensive study on different sizes from the trained objects shows that our model can generalize well to different size objects, especially smaller objects, due to more available picking actions in less crowded spaces.

\subsubsection{Unseen Shape}
\label{diff_shape_section}

\begin{table}[h!]
\centering
\caption{Hexagon Table}
\label{extensive_diff_shape}
\begin{tabular}{|c|c|c|c|} 
\hline
\multicolumn{3}{|c|}{}                                                                                                                                           & Target   \\ 
\hline
Setting                                                                    & Obj Index                                                                  & Result & 2        \\ 
\hline
\multirow{6}{*}{\begin{tabular}[c]{@{}c@{}}Simulation\\(200)\end{tabular}} & \multirow{2}{*}{\begin{tabular}[c]{@{}c@{}}hexa\_s\_s \\(20)\end{tabular}} & OSR    & 62.50\%  \\ 
\cline{3-4}
                                                                           &                                                                            & ESR    & 62.50\%  \\ 
\cline{2-4}
                                                                           & \multirow{2}{*}{\begin{tabular}[c]{@{}c@{}}hexa\_m\_s \\(20)\end{tabular}} & OSR    & 67.00\%  \\ 
\cline{3-4}
                                                                           &                                                                            & ESR    & 67.00\%  \\ 
\cline{2-4}
                                                                           & \multirow{2}{*}{\begin{tabular}[c]{@{}c@{}}hexa\_l\_s\\(20)\end{tabular}}  & OSR    & 61.00\%  \\ 
\cline{3-4}
                                                                           &                                                                            & ESR    & 61.00\%  \\ 
\hline
\multirow{2}{*}{\begin{tabular}[c]{@{}c@{}}Real\\(80)\end{tabular}}        & \multirow{2}{*}{\begin{tabular}[c]{@{}c@{}}hexa\_m\_s\\(15)\end{tabular}}   & OSR    & 63.75\%  \\ 
\cline{3-4}
                                                                           &                                                                            & ESR    & 65.38\%  \\
\hline
\end{tabular}
\end{table}

We also evaluated our approach to an unseen shape, a hexagon, in both simulation and real testing. Our shorter gripper MOP predictor has never seen a hexagon shape during the training stage. Therefore this testing can give us insight into the possibility our model can generalize to more complex shapes in future works. Table \ref{extensive_diff_shape} shows the picking result of the hexagon in three different sizes in simulation and metal hexagonal nuts. The result shows $hexa\_m\_s$ has the highest OSR and ESR in simulation when picking two objects, reaching 67.00\% in both success rates. We believe it is because this hexagon's size is closest to the trained 1inch cube $cube\_m\_s$.

In real-world testing, the hexagonal metal nuts have the same size as $hexa\_m\_s$. The result OSR reaches 63.75\% and ESR reaches 65.38\%. Therefore, our trained model and algorithm can partially generalize to a shape never seen before.

Our algorithm also shows the trained model can generalize to different size cube and cylinder: 2.0cm and 3.0cm cubes achieved more than 90\% overall accuracy when picking 2 objects. The result shows the model can generalize to both size, but the overall success rate for larger than original object decreases due to more collision. The evaluation result also shows that hexagon can achieves 67.00\% and 63.75\% overall success rate for target\#=2 picking in simulation and real. This success rate is lower than original trained shape, but still better than SOP theoretical result.

\section{Conclusion and Discussion}
\label{conclusion_and_discussion}

The proposed OPOS can use a robotic gripper to pick up the requested number of objects from a shallow bin by only picking once with a reasonable success rate. Using OPOS can improve picking efficiency even when considering scenes that are not naturally suitable for OPO. OPOS has a novel framework building on several graph algorithms that use the graph topology of object layouts to identify suitable clusters and a novel neural-network MOP predictor to estimate the outcome of a picking from a proposed gripper pose. For complex scenes with many objects, the framework has a cluster ranking algorithm and an early-stopping mechanism to reduce the computation cost significantly. 

The approach has been evaluated with 12 objects in the simulation and 9 objects in the real setup through three protocols on four metrics. The results show when the requested number is two, our OPOS achieves close to or above 95\% overall OPO success rates for all tested objects in various shapes and sizes in the simulation and real when the density is not too high. When enforcing the requested number, OPOS may have to pick more than once to reach the exact requested number. The results show the OPOS approach can improve efficiency by almost three times for some objects and settings and close to two times for objects and all settings. The ablation study shows our algorithms achieve the best overall performance in terms of planning efficiency and overall success rate. The generalization results show the trained model can generalize to slightly different sizes and shapes without any training. Their OPO success rates and efficiencies decrease a bit, but they are still better than a perfect SOP.

Picking a large (above three) requested number of objects by OPO is usually difficult because of lack of clusters and the high chance of collision (a large number requires a long gripper that has a high chance of collision). For identical settings (the same object, the same density), the AR decreases as the target number increases. The AR significantly increases when density increases for the same object and the same requested number of objects (especially a high number). It is because higher density is more likely to provide clusters of a larger number of objects to be picked up together. 

Overall, the proposed OPOS can improve the efficiency of picking multiple identical objects. When implementing OPOS for a particular setup, the requested number could be broken down to two or three for each OPOS picking to achieve the best efficiency. OPOS could be improved if we introduce a gathering motion because it will increase the OPO availability rate. In the future, we plan to investigate different gathering motions and how to incorporate them into OPOS. 

\bibliographystyle{IEEEtran}

\bibliography{references}

\end{document}